\documentclass[sigconf]{acmart}

\AtBeginDocument{%
  \providecommand\BibTeX{{%
    \normalfont B\kern-0.5em{\scshape i\kern-0.25em b}\kern-0.8em\TeX}}}

\copyrightyear{2022}
\acmYear{2022}
\setcopyright{none}
\acmConference[WWW '22] {Proceedings of the ACM Web Conference 2022}{April 25--29, 2022}{Virtual Event, Lyon, France.}
\acmBooktitle{Proceedings of the ACM Web Conference 2022 (WWW '22), April 25--29, 2022, Virtual Event, Lyon, France}
\acmPrice{15.00}
\acmISBN{978-1-4503-9096-5/22/04}
\acmDOI{10.1145/3485447.3512008}

\usepackage{array}
\usepackage{times}  
\usepackage{helvet}  
\usepackage{courier}  
\usepackage{algorithm}
\usepackage{algorithmic}

\usepackage{amsmath}
\usepackage{amsthm}
\usepackage{enumerate}
\usepackage{subfigure}
\usepackage{bm}
\usepackage{tabularx}
\usepackage{enumitem}
\usepackage{multirow}
\usepackage{xcolor}
\usepackage{nicefrac}
\usepackage{booktabs}

\newtheorem{theorem}{Theorem}

\newtheorem{prop}{Proposition}

\newtheorem{definition}{Definition}

\newcolumntype{L}[1]{>{\raggedright\let\newline\\\arraybackslash\hspace{0pt}}m{#1}}
\newcolumntype{C}[1]{>{\centering\let\newline  \\\arraybackslash\hspace{0pt}}m{#1}}%
\newcolumntype{R}[1]{>{\raggedleft\let\newline \\\arraybackslash\hspace{0pt}}m{#1}}

\begin{document}

\title{Knowledge Graph Reasoning with Relational Digraph}

\author{Yongqi Zhang
}
\affiliation{%
	\institution{4Paradigm Inc.}
	\city{Beijing}
	\country{China}}
\email{zhangyongqi@4paradigm.com}

\author{Quanming Yao}
\affiliation{
	\institution{EE, Tsinghua University}
	\city{Beijing}
	\country{China}
}
\email{qyaoaa@tsinghua.edu.cn}

%


\begin{abstract}
Reasoning on the knowledge graph (KG)
aims to infer new facts from existing ones.
Methods based on the relational path 
have shown strong, interpretable,
and transferable reasoning ability.
However,                                              
paths are naturally limited in capturing local evidence in graphs.
In this paper,
we introduce a novel relational structure,
i.e., relational directed graph (r-digraph),      
which is composed of overlapped relational paths,
to capture the KG's local evidence.
Since the r-digraphs are more complex than paths,
how to efficiently construct and effectively learn from them are challenging.
Directly encoding the r-digraphs
cannot scale well
and capturing query-dependent information is hard in r-digraphs.
We propose a variant of graph neural network, 
i.e.,
RED-GNN,
to address the above challenges.
Specifically,
RED-GNN makes use of dynamic programming to
recursively encodes multiple r-digraphs with shared edges,
and utilizes query-dependent attention mechanism to
select the strongly correlated edges.
We demonstrate that
RED-GNN 
is not only efficient 
but also 
can achieve 
significant performance gains 
in both inductive and transductive reasoning tasks
over existing methods.
Besides,
the learned attention weights in
RED-GNN can exhibit 
interpretable evidence for KG reasoning.
\footnote{The code is available at \url{https://github.com/AutoML-Research/RED-GNN/}. 
	Correspondence is to Q. Yao.}
\end{abstract}


\maketitle

\section{Introduction}
\label{sec:intro}

Knowledge graph (KG),
which contains the interactions among real-world objects, peoples, concepts, etc.,
brings much connection between artificial intelligence and human knowledge~\cite{battaglia2018relational,ji2020survey,hogan2021knowledge}.
The interactions are represented as facts 
with the triple form
\textit{(subject entity, relation, object entity)}
to indicate the relation between entities.
The real-world KGs are large and highly incomplete~\cite{wang2017knowledge,ji2020survey},
thus inferring new facts is challenging.
KG reasoning simulates such a process to deduce new facts from existing facts~\cite{chen2020review}.
It has wide application in 
semantic search~\cite{berant2014semantic},
recommendation~\cite{cao2019unifying},
and
question answering~\cite{abujabal2018never}, etc.
In this paper,
we focus on learning the relational structures for reasoning 
on
the queries 
in the form of \textit{(subject entity, relation, ?)}.

Over the last decade,
triple-based models have gained much attention to 
learn semantic information in KGs~\cite{wang2017knowledge}.
These models 
directly 
reason on triples
with 
the entity and relation
embeddings,
such as
TransE~\cite{bordes2013translating},
ConvE~\cite{dettmers2017convolutional},
ComplEx~\cite{trouillon2017knowledge},
RotatE~\cite{sun2019rotate}, 
QuatE~\cite{zhang2019quaternion}, 
AutoSF~\cite{zhang2020autosf},
etc.
Since 
the triples are independently learned,
they cannot explicitly capture the 
\textit{local evidence}~\cite{niepert2016discriminative,xiong2017deeppath,yang2017differentiable,teru2019inductive},
i.e., the local structures around the query triples,
which can be used as evidence for KG reasoning~\cite{niepert2016discriminative}.

Learning on paths
can help to better capture local evidences in graphs 
since they can preserve sequential connections between nodes~\cite{perozzi2014deepwalk,grover2016node2vec}.
Relational path
is the first attempt to capture both the semantic and local evidence for reasoning~\cite{lao2010relational}.
DeepPath~\cite{xiong2017deeppath}, 
MINERVA~\cite{das2017go} and M-walk~\cite{shen2018m}
use reinforcement learning (RL) to
sample
relational paths that have strong correlation with the queries.
Due to the sparse property of KGs,
the RL approaches are hard to train on large-scale KGs~\cite{chen2020review}.
PathCon~\cite{wang2021relational}
samples all the relational paths between the entities
and use attention mechanism to weight the different paths,
but is expensive for the entity query tasks.
The rule-based methods,
such as RuleN~\cite{meilicke2018fine}, 
Neural LP~\cite{yang2017differentiable},
DRUM~\cite{sadeghian2019drum}
and RNNLogic~\cite{qu2021rnnlogic},
generalize the relational paths
as logical rules,
which learn to infer relations by logical composition of relations, 
and can provide interpretable insights.
Besides,
the logical rules can 
handle inductive reasoning
where there exist unseen entities in the inference,
which are common in real-world applications~\cite{yang2017differentiable,sadeghian2019drum,zhang2019inductive}. 

Subgraphs can naturally be more informative than paths
in capturing the local evidence~\cite{battaglia2018relational,ding2021diffmg,xiao2021neural}.
Their effectiveness has been empirically
verified in, e.g., 
graph-based recommendation~\cite{zhang2019inductive,zhao2017meta}
and node representation learning~\cite{hamilton2017inductive}.
With the success of 
graph neural network (GNN)~\cite{gilmer2017neural,kipf2016semi}
in modeling graph-structured data,
GNN has been introduced to capture the subgraph structures in KG.
R-GCN~\cite{schlichtkrull2018modeling},
CompGCN~\cite{vashishth2019composition} 
and KE-GCN~\cite{yu2021knowledge}
propose
to update the representations of entities by aggregating all 
the neighbors in each layer.
However,
they cannot distinguish the structural role of different neighbors
and cannot be interpretable.
DPMPN~\cite{xu2019dynamically} learns to reduce the size of subgraph
for reasoning on large-scale KGs
by pruning the irrelevant entities for a given query
rather than learning the specific local structures.
p-GAT \cite{harsha2020probabilistic}
jointly learns a Graph Attention Network
and a Markov Logic Network
to bridge the gap between embeddings and rules.
However,
a set of rules must be pre-defined
and 
the expensive EM-algorithm should be used for optimization.
Recently,
GraIL~\cite{teru2019inductive}
proposes to predict relation from the local enclosing subgraph                 
and shows the inductive ability of subgraph.
However,
it suffers from both effectiveness and efficiency problems
due to the limitation of the enclosing subgraph.

Inspired by the interpretable and transferable abilities of path-based methods
and the structure preserving property of subgraphs,
we introduce a novel relational structure into KG, 
called r-digraph,
to combine the best of both worlds.
The r-digraphs generalize relational paths to subgraphs
by preserving the overlapped relational paths and the structures of relations for reasoning.
Different from the relational paths that have simple structures,
how to efficiently construct and learn from the r-digraphs are challenging
since 
the construction process is expensive \cite{teru2019inductive,wang2021relational}.
Inspired by saving computation costs in overlapping sub-problems using dynamic programming,
we propose RED-GNN,
a variant of GNN \cite{gilmer2017neural},
to recursively encode multiple \textbf{RE}lational \textbf{D}igraphs (r-digraphs)
with shared edges
and select the strongly correlated edges through  query-dependent attention weights.
Empirically,
RED-GNN shows significant gains over the state-of-the-art reasoning methods in both 
inductive and transductive benchmarks.
Besides,
the training and inference processes are efficient,
the number of model parameters are small,
and the learned structures are interpretable.

\section{Related Works}
\label{sec:related}

A knowledge graph is in the form of $\mathcal K\!=\!\{\mathcal V, \mathcal R, \mathcal F \}$,
where $\mathcal V$, $\mathcal R$ and $\mathcal F\!=\!\{(e_s,r,e_o)|e_s,e_o\!\in\!\mathcal V, r\!\in\!\mathcal R\}$
are the sets of entities, relations and fact triples, respectively.
Let $e_q$ be the query entity, $r_q$ be query relation, and $e_a$ be answer entity.
Given the query $(e_q, r_q, ?)$,
the reasoning task is to predict the missing answer entity $e_a$.
Generally,
all the entities in $\mathcal V$ are candidates for $e_a$ 
\cite{chen2020review,das2017go,yang2017differentiable}.

The key for KG reasoning is to capture 
the \textit{local evidence},
i.e.,
entities and relations around the query triples $(e_q, r_q, e_a)$.
We regard $(e_q, r_q, e_a)$ as missing link,
thus the local evidence does not contain this triplet in.
Examples of such local evidence explored in the literature are
relational paths \cite{das2017go,lao2011random,xiong2017deeppath,wang2021relational}
and 
subgraphs \cite{schlichtkrull2018modeling,teru2019inductive,vashishth2019composition}.
In this part,
we introduce 
the path-based methods and GNN-based methods that
leverage structures in $\mathcal F$ for reasoning.

\subsection{Path-based methods}

Relational path (Definition~\ref{def:path}) is a set of triples
that are sequentially connected.
The path-based methods learn to predict the triple $(e_q, r_q, e_a)$
by a set of relational paths
as local evidence.
DeepPath~\cite{xiong2017deeppath}
learns to generate the relational path from $e_q$ to $e_a$ by reinforcement learning (RL).
To improve the efficiency,
MINERVA~\cite{das2017go} and M-walk~\cite{shen2018m} learn to
generate multiple paths starting from 
$e_q$ by RL.
The scores are indicated by the arrival frequency on different $e_a$'s.
Due to the complex structure of KG,
the reward is very sparse,
making it hard to train the RL models~\cite{chen2020review}.
PathCon~\cite{wang2021relational}
samples all the paths connecting the two entities
to predict the relation between them,
which is expensive for the reasoning task
$(e_q, r_q, e_a)$.

\begin{definition}[Relational path~\cite{lao2011random,xiong2017deeppath,zhang2020interstellar}]
	\label{def:path}
	The relational path with length $L$ is a set of $L$ triples
	$(e_0, r_1, e_1)$, 
	$(e_1, r_2, e_2)$, 
	$\dots$, 
	$(e_{L-1}, r_L, e_L)$,
	that are connected head-to-tail sequentially.
\end{definition}

Instead of directly using paths,
the rule-based methods learn logical rules
as a generalized form of relational paths.
The logical rules are formed with the composition of a set of relations to infer a specific relation.
This can provide better interpretation and can be transfered to unseen entities. 
The rules are learned
by either the mining methods like RuleN~\cite{meilicke2018fine},
EM algorithm like RNNLogic~\cite{qu2021rnnlogic},
or end-to-end training,
such as 
Neural LP~\cite{yang2017differentiable} and DRUM~\cite{sadeghian2019drum},
to generate highly correlated relational paths between $e_q$ and $e_a$.
The rules can provide logical interpretation and transfer to unseen entities.
However,
the rules only capture the sequential evidences,
thus cannot learn more complex patterns
such as the 
subgraph structures.

\begin{figure*}[t]
	\centering
	\subfigure[Knowledge graph.]
	{\includegraphics[height=3.5cm]{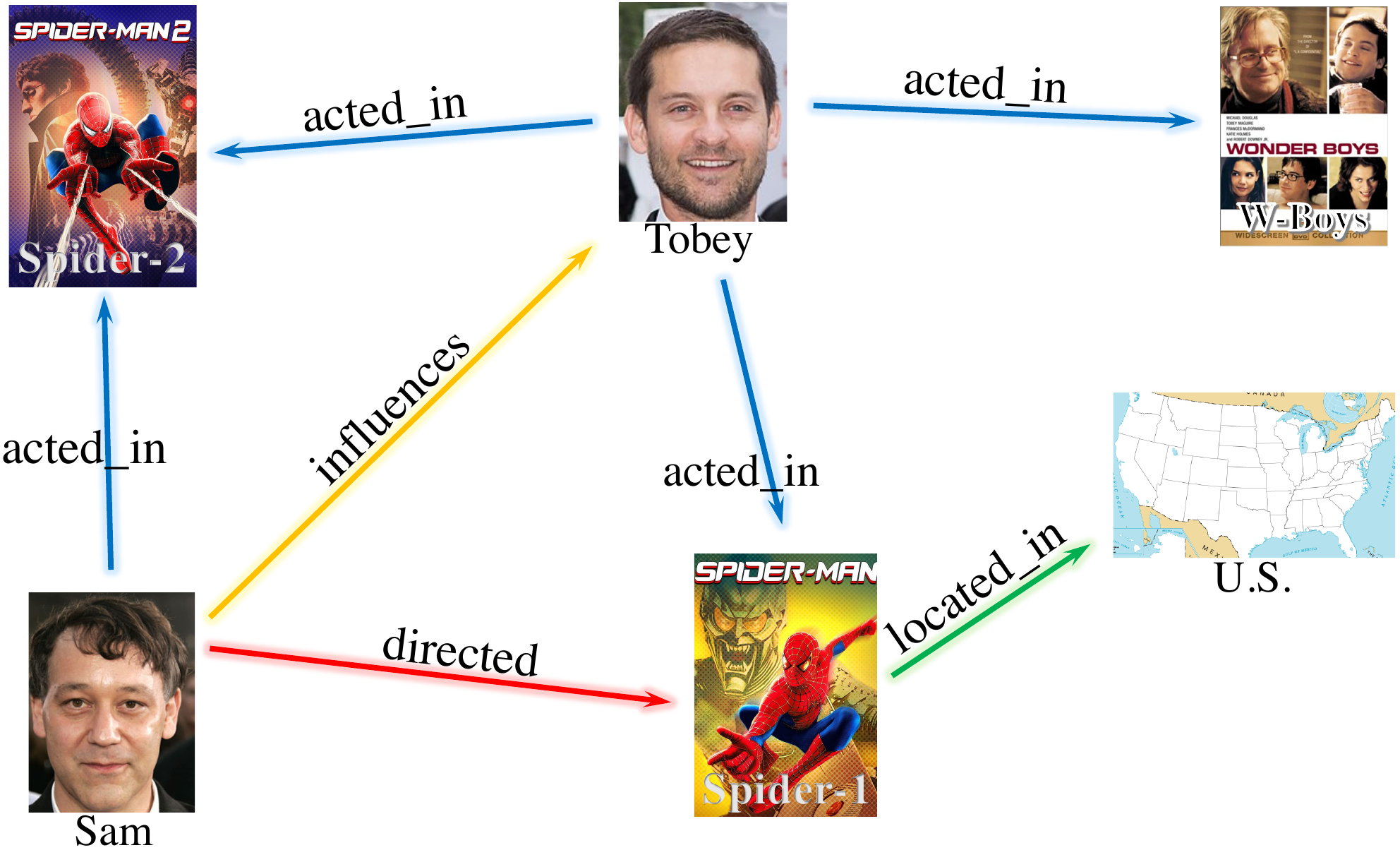}  \label{fig:KG}}
	\quad\ 
	\subfigure[$\mathcal G_{\text{Sam,Spider-2}|3}$.]
	{\includegraphics[height=3.5cm]{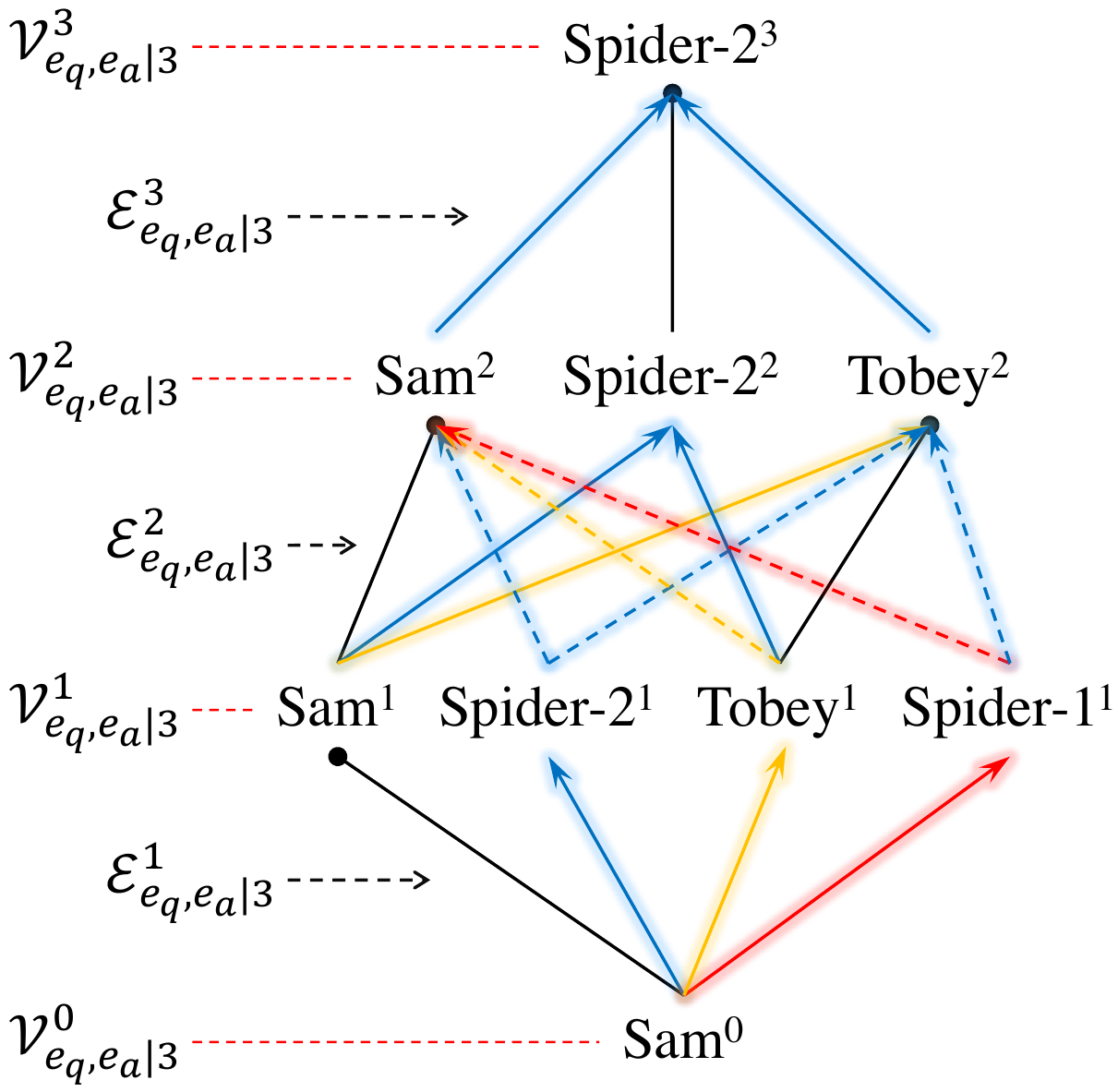} \label{fig:relsub}}
	\quad\ 
	\subfigure[Recursive encoding.]
	{\includegraphics[height=3.5cm]{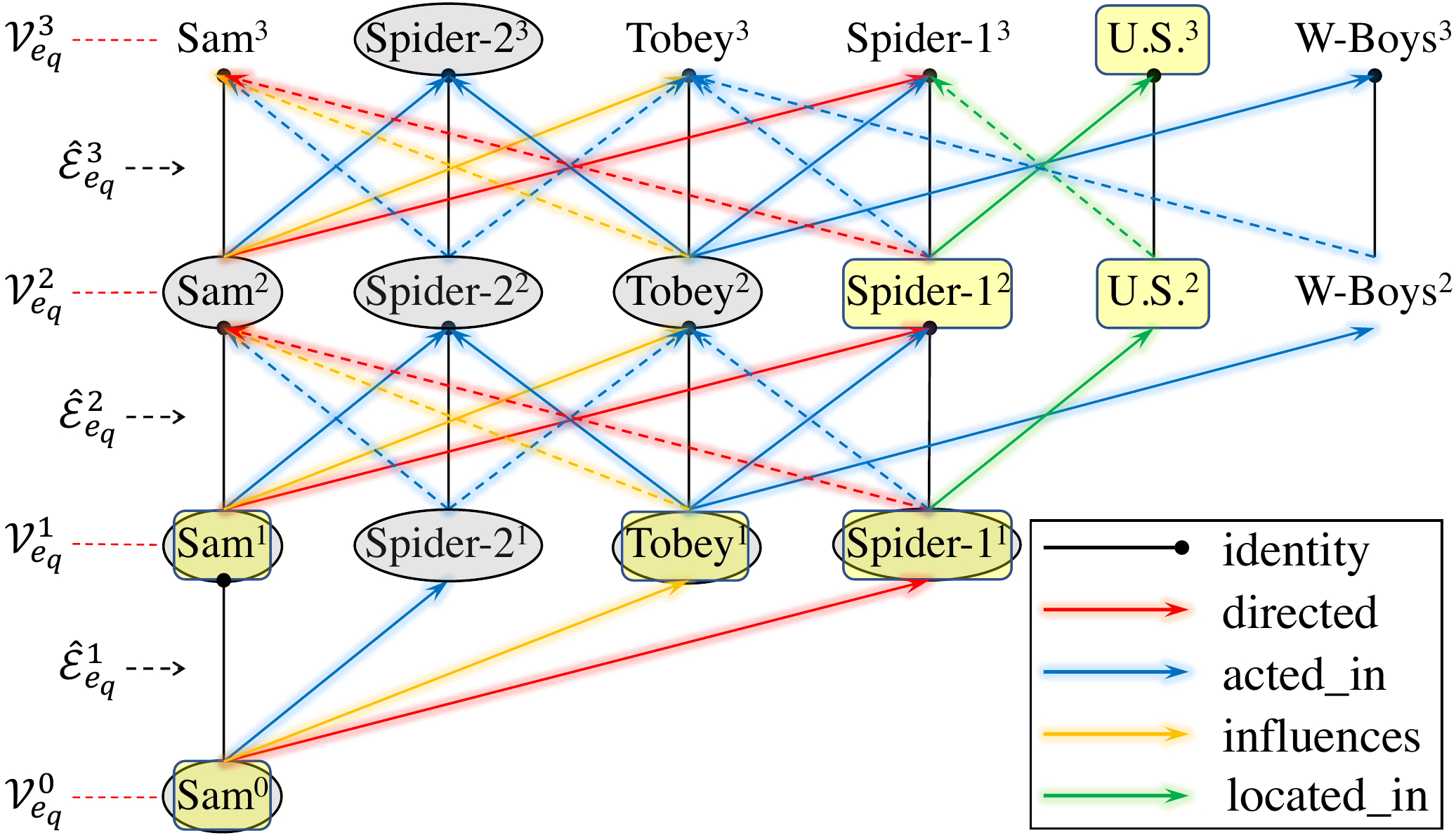} \label{fig:forward}}
	\vspace{-7px}
	\caption{
		Graphical illustration.
		In (c), 
		the subgraph formed by the gray ellipses is $\mathcal G_{\emph{Sam, Spider-2}|3}$
		and the subgraph formed by the yellow rectangles is $\mathcal G_{\emph{Sam,U.S.}|3}$.
		Dashed edges mean the reverse relations of corresponding color (best viewed in color).}
	\label{fig:graphical}
\end{figure*}

\subsection{GNN-based methods}
\label{ssec:related-gnn}

As mentioned in Section~\ref{sec:intro},
the subgraphs can preserve richer information than the relational paths
as 
more degree are allowed in the subgraph.
GNN
has shown strong power in modeling the graph structured data~\cite{battaglia2018relational}.
This inspires recent works,
such as R-GCN~\cite{schlichtkrull2018modeling}, 
CompGCN~\cite{vashishth2019composition},
and KE-GCN~\cite{yu2021knowledge},
extending GNN on KG to aggregate the entities' and relations' representations
under the message passing framework~\cite{gilmer2017neural} as
\begin{align}
\bm{h}_{e_o}^{\ell} &= \delta\Big(\bm{W}^{\ell} \cdot \sum\nolimits_{(e_s, r, e_o)\in\mathcal F} \phi\big(\bm h^{\ell-1}_{e_s}, \bm h_r^\ell\big)\Big),
\label{eq:kggnn}
\end{align}
which aggregates the message $\phi(\cdot, \!\cdot)$ on the $1$-hop neighbor edges $(e_s, r, e_o)\!\in\!\mathcal F$ of entity $e_o$
with dimension $d$.
$\bm W^\ell\!\!\in\!\mathbb R^{d\times d}$ is a weighting matrix,
$\delta$ is the activation function
and $\bm h_r^\ell$ is the relation representation
in the $\ell$-th layer.
After $L$ layers'
aggregation,
the representations $\bm h^L_e\!$
capturing the local structures of entities $e\!\in\!\mathcal V$
jointly work with a decoder scoring function to measure the triples.
Since the 
message passing function \eqref{eq:kggnn} aggregates information of all the neighbors
and is independent with the query,
R-GCN and CompGCN cannot capture the explicit local evidence for specific queries
and are not interpretable.

Instead of using all the neighborhoods,
DPMPN~\cite{xu2019dynamically}
designs one GNN to aggregate the
embeddings of entities,
and another GNN to dynamically 
expand and prune the inference subgraph from the query entity $e_q$.
A query-dependent attention is applied over the sampled entities for pruning.
This approach shows interpretable reasoning process 
by the attention flow on the pruned subgraph,
but 
still requires embeddings to guide the pruning,
thus cannot be generalized to unseen entities.
Besides,
it cannot capture the explicit local evidence supporting a given query triple.

p-GAT~\cite{harsha2020probabilistic} and pLogicNet~\cite{qu2019probabilistic}
jointly learns an embedding-based model
and a Markov logic network (MLN) by variational-EM algorithm.
The embedding model generates evidence for MLN to update
and MLN expends the training data for embedding model to train.
MLN brings the gap between embedding models and logic rules,
but it 
requires a pre-defined set of rules for initialization
and the training is expensive with the variational-EM algorithm.

Recently,
GraIL~\cite{teru2019inductive} proposes to extract
the enclosing subgraph $\mathcal G_{(e_q, e_a)}$
between the query entity $e_q$ and answer entity $e_a$.
To learn the enclosing subgraph,
the relational GNN~\cite{schlichtkrull2018modeling}
with
query-dependent attention is applied over the edges in $\mathcal G_{(e_q, e_a)}$
to control the importance of edges for different queries,
but the learned attention weights are not interpretable (see Appendix~\ref{app:grail}).
After $L$ layers' aggregation,
the graph-level representations aggregate
all the entities $e\in\mathcal V$ in the subgraph 
and are used to score the triple $(e_q, r_q, e_a)$.
Since the subgraphs need to be explicitly extracted and scored for different triples,
the computation cost is very high.

\section{Relational Digraph (r-digraph)}

The relational paths
have shown strong transferable
and interpretable reasoning ability on KGs~\cite{das2017go,yang2017differentiable,sadeghian2019drum}.
However,
they are limited in capturing more complex dependencies in KG 
since the nodes in paths are only sequentially connected.
GNN-based methods can learn different subgraph structures.
But none of the existing methods can efficiently learn  
the subgraph structures
that are both interpretable and inductive like rules.
Hence,
we are motivated to define a new kind of structure, 
to explore the important local evidence.

Before defining r-digraph,
we first introduce a special type of directed graph in Definition~\ref{def:stdag}.

\begin{definition}[Layered \textit{st}-graph~\cite{battista1998graph}]
	\label{def:stdag}
	The layered \textit{st}-graph is a directed graph with exactly one source node (s) and one sink node (t).
	All the edges are directed, connecting nodes between consecutive layers
	and pointing from $l$-th layer to $l+1$-th layer.
\end{definition}

Here,
we adopt the general approaches to augment the triples
with reverse and identity relations~\cite{vashishth2019composition,sadeghian2019drum}.
Then, 
all the relational paths 
with length less than or equal to $L$ between $e_q$ and $e_a$
can be represented as relational paths
$e_q \!\rightarrow_{r^1} \!\cdot\!\rightarrow_{r^2}\! \cdots \!\rightarrow_{r^L} \!\!e_a$
with length $L$.
In this way,
they
can be formed 
as paths in
a layered \textit{st}-graph,
with the single source entity $e_q$ and sink entity $e_a$.
Such a structure 
preserves all the relational paths 
between $e_q$ and $e_a$ up to length $L$, 
and maintains the subgraph structures.
Based on this observation,
we define r-digraph in Definition~\ref{def:relgraph}.

\begin{definition}[r-digraph]
	\label{def:relgraph}
	The r-digraph $\mathcal G_{e_q, e_a|L}$
	is a layered \textit{st}-graph
	with the source entity $e_q$ and the sink entity $e_a$.
	The entities in the same layer are different with each other.
	Any path pointing from $e_q$ to $e_a$ in the r-digraph
	is a relational path $e_q\!\rightarrow_{r^1}\!\cdot\!\rightarrow_{r^2}\!\cdots \!\rightarrow_{r^L}\!e_a$ with length $L$,
	where $r^{\ell}$ 
	connects an entity in the $\ell\!\!-\!\!1$-layer to an entity in $\ell$-layer.
	We define
	$\mathcal G_{e_q, e_a|L}=\emptyset$ if there is no relational path connecting $e_q$ and $e_a$ with length $L$.
\end{definition}

Figure~\ref{fig:relsub} provides an example of r-digraph $\mathcal G_{\text{Sam,Spider-2}|3}$,
which is used to infer the new triple \textit{(Sam, directed, Spider-2)},
in the exemplar KG in Figure~\ref{fig:KG}.
Inspired by the reasoning ability of relational paths 
\cite{das2017go,yang2017differentiable,sadeghian2019drum},
we aim to leverage the r-digraph
for KG reasoning.
However,
different from the relational paths
which have simple structures to learn with sequential models \cite{xiong2017deeppath,das2017go},
how to efficiently construct and how to effectively learn from the r-digraphs are challenging.

In the paper,
$\mathcal E_{e_q, e_a|L}^\ell$ 
are the edges 
and 
$\mathcal V_{e_q, e_a|L}^\ell
= \big\{ e_o|(e_s, r, e_o)\!\!\in\!\! \mathcal E_{e_q, e_a|L}^\ell \big\}$ are the entities
in the $\ell$-th layer of the r-digraph
\vspace{-5px}
\begin{equation*}
\mathcal G_{e_q, e_a|L} = \mathcal E_{e_q, e_a|L}^1\otimes \cdots \otimes \mathcal E_{e_q, e_a|L}^L
\;,
\end{equation*}
with
$\otimes$ denoting the layer-wise connection.
We define the union operator as
$\mathcal G_{e_{q_1}, e_{a_1}|L}\!\cup\! \mathcal G_{e_{q_2}, e_{a_2}\!|L} 
\!=\! \big(\mathcal E_{e_{q_1}, e_{a_1}\!|L}^1\!\cup\mathcal E_{e_{q_2}, e_{a_2}\!|L}^1\big) 
\!\otimes \!\cdots \!\otimes \! 
\big(
\mathcal E_{e_{q_1}, e_{a_1}\!|L}^L\!\cup\mathcal E_{e_{q_2}, e_{a_2}\!|L}^L
\big)$.
Given an entity $e$,
we denote $\hat{\mathcal E}_{e}^\ell$, $\check{\mathcal E}_{e}^\ell$ and ${\mathcal V}_{e}^\ell$
as the sets of out-edges,
in-edges
and entities, respectively,
visible in $\ell$ steps walking {from} $e$.
$\mathcal E_{e_q, e_a|L}^\ell, \mathcal V_{e_q, e_a|L}^\ell$
and
$\hat{\mathcal E}_{e_q}^\ell, {\mathcal V}_{e_q}^\ell$
are graphically shown in Figure~\ref{fig:graphical}.

\section{The Proposed Model}
\label{sec:relgnn}

Here,
we show how GNNs can be tailor-made to efficiently and effectively learn from 
the r-digraphs.
Extracting subgraph structures and
then learning the subgraph representation
is a common practice
for subgraph encoding in the literature,
such as 
GraphSage \cite{hamilton2017inductive}
and GraIL \cite{teru2019inductive}.
Given a query triple $(e_q, r_q, e_a)$,
the subgraph encoding generally contains three processes:
\begin{enumerate}[leftmargin=20pt]
	\item[(i).] extract the neighborhoods of both $e_q$ and $e_a$;
	\item[(ii).] take the intersection to construct the subgraph;
	\item[(iii).]  run message passing and use the graph-level representation
	as the subgraph encoding.
\end{enumerate}
When working on the r-digraph $\mathcal G_{e_q, e_a|L}$,
the same approach can be customized in Algorithm~\ref{alg:simple}.
First,
we get the neighborhoods of both $e_q$ and $e_a$
in steps~\ref{step:neib1}-\ref{step:neib2}.
Second,
we take the intersection of neighborhoods from $e_q$ and $e_a$
in steps \ref{step:neib3}-\ref{step:neib4}
to induce the r-digraph $\mathcal G_{e_q, e_a|L}$ layer-wisely.
Third,
if the r-digraph is empty, we set the representation as $\bm 0$ in step~\ref{step:empty}.
Otherwise,
the message passing is conducted layer-by-layer in steps~\ref{step:mp:s}-\ref{step:mp:e}.
Since $e_a$ is the single sink entity,
the final layer representation $\bm h_{e_a}^L\!(e_q, r_q)$
is used as subgraph representation to encode the r-digraph $\mathcal G_{e_q, e_a|L}$.
We name this simple solution as \textit{RED-Simp}.

\begin{algorithm}[ht]
	\caption{RED-Simp: Message passing on single r-digraph}
	\small
	\label{alg:simple}
	\begin{algorithmic}[1]
		\STATE initialize $\bm{h}_{e_q}^{0}\!(e_q, r_q)\!=\!\bm0$ and the entity sets $\mathcal V_{e_q}^0 \!=\! \{e_q\}$, $\mathcal V_{e_a}^0 \!=\! \{e_a\}$;
		\FOR{$\ell=1,2,\dots, L$}  \label{step:neib1}
		\STATE get the $\ell$-hop out-edges $\hat{\mathcal E}_{e_q}^{\ell}\!=\!\{(e_s, r, e_o)\!\in\!\mathcal F| e_s\!\in\!\mathcal V_{e_q}^{\ell-1}\}$   \\  \label{step:eq}
		and  entities $\mathcal V_{e_q}^{\ell} \!=\! \{e_o| (e_s, r, e_o)\in \hat{\mathcal E}_{e_q}^{\ell}\!\}$ of $e_q$;
		\STATE get the $\ell$-hop in-edges $\check{\mathcal E}_{e_a}^{\ell}\!=\!\{(e_s, r, e_o)\!\in\!\mathcal F| e_o\!\in\!\mathcal V_{e_a}^{\ell-1}\}$   \\ \label{step:ea}
		and  entities $\mathcal V_{e_a}^{\ell} \!=\! \{e_s| (e_s, r, e_o)\in \check{\mathcal E}_{e_a}^{\ell}\!\}$ of $e_a$;
		\ENDFOR   \label{step:neib2}
		
		\FOR{$\ell=0,1,\dots, L$}  \label{step:neib3}
			\STATE intersection 
			$\mathcal E_{e_q, e_a|L}^\ell \!=\! \hat{\mathcal E}_{e_q}^{\ell} \!\cap\! \check{\mathcal E}_{e_a}^{L-\ell}$
			and
			$\mathcal V_{e_q, e_a|L}^\ell \!=\! \mathcal V_{e_q}^{\ell}\!\cap\! \mathcal V_{e_a}^{L-\ell}$;
		\ENDFOR  \label{step:neib4}
		
		\STATE \textbf{if} $\mathcal V_{e_q, e_a|L}^L=\emptyset~~~~$ \textbf{return} $\bm h_{e_a}^L(e_q, r_q) = \bm 0$; \label{step:empty}
		\FOR{$\ell=1,2,\dots, L$} \label{step:mp:s}
		\STATE message passing for entities $e_o\!\in\!\mathcal V_{e_q, e_a|L}^\ell$: \\$\bm{h}_{e_o}^{\ell}(e_q,r_q) = \delta
		\Big(\bm{W}^{\ell} \!\cdot\! \sum\nolimits_{(e_s, r, e_o)\in\mathcal E_{e_q, e_a|L}^\ell} \phi\big(\bm h^{\ell-1}_{e_s}\!(e_q,r_q), \bm h_r^\ell\big)\Big)$;  
		\ENDFOR \label{step:mp:e}
		\RETURN $\bm h_{e_a}^L\!(e_q, r_q)$.
	\end{algorithmic}
\end{algorithm}

However,
Algorithm~\ref{alg:simple} is very expensive.
First,
we need to conduct the two directional sampling in steps~\ref{step:eq}
and \ref{step:ea},
and take the intersection to extract the r-digraph.
Second,
given a query $(e_q, r_q, ?)$,
we need to do this algorithm for $|\mathcal V|$ different triples
with different answering entities $e_a\!\in\!\mathcal V$.
It needs $O\big(|\mathcal V|\cdot(\min(\bar{D}^L, |\mathcal F|L)+d\bar{E}L)\big)$
time 
(see Section~\ref{ssec:complexity})
to predict a given query $(e_q, r_q, ?)$,
where 
$\bar{D}$ is the average degree of entities in $\mathcal V$
and
$\bar{E}$ is the average number of edges in $\mathcal E_{e_q, e_a|L}^\ell$'s.
These limitations also exist in PathCon~\cite{wang2021relational} and GraIL~\cite{teru2019inductive}.
To improve efficiency,
we propose to encode multiple r-digraphs recursively in Section \ref{ssec:subgraph}.


\subsection{Recursive r-digraph encoding}
\label{ssec:subgraph}

In Algorithm~\ref{alg:simple},
when evaluating 
$(e_q, r_q, e_a)$ with different $e_a \!\in\! \mathcal{V}$ but the same query $(e_q, r_q, ?)$,
the neighboring edges $\hat{\mathcal E}_{e_q}^{\ell}, \ell\!=\!1\!\dots\! L$ of $e_q$ are shared.
We have the following observation:

\begin{prop}
	\label{pr:eqneighbor}
	The set of
	edges $\hat{\mathcal E}_{e_q}^\ell$ visible from $e_q$ by $\ell$-steps 
	equals $\cup_{e_a\in\mathcal V}\mathcal E_{e_q, e_a|L}^\ell$,
	namely
	 $\hat{\mathcal E}_{e_q}^\ell$ is the union
	of the $\ell$-th layer edges in the r-digraphs between $e_q$ and all the entities $e_a\in\mathcal V$.
\end{prop}

Proposition~\ref{pr:eqneighbor} indicates that 
the $\ell$-th layer edges $\mathcal E_{e_q, e_a|L}^\ell$ with different answer entities $e_a$ 
share the same set of edges in $\hat{\mathcal E}_{e_q}^\ell$.
A common approach to save computation cost in overlapping sub-problems
is dynamic programming.
This has been used to aggregate node representations on large-scale graphs \cite{hamilton2017inductive}
or to propagate the question representation on KGs \cite{zhang2018variational}.
Inspired by the efficiency gained by dynamic programming,
we recursively construct
the r-digraph between $e_q$ and any entity $e_o$ as
\begin{equation}
\mathcal G_{e_q, e_o|\ell} 
= \cup_{(e_s, r, e_o)
\in
\hat{\mathcal E}_{e_q}^\ell}\mathcal G_{e_q, e_s|\ell-1}
\otimes
\left\lbrace 
(e_s, r, e_o)\in\hat{\mathcal E}_{e_q}^\ell
\right\rbrace .
\label{eq:dm}
\end{equation}
Once the representations of $\mathcal G_{e_q, e_s|\ell-1}$ for all the entities $e_s\in\mathcal V_{e_q}^{\ell-1}$ in the $\ell\!-\!1$-th layer are ready,
we can encode $\mathcal G_{e_q, e_o|\ell}$ by combining $\mathcal G_{e_q, e_s|\ell-1}$ 
with the shared edges 
$(e_s, r, e_o) \!\in\! \hat{\mathcal E}_{e_q}^\ell$ in the $\ell$-th layer.
Based on Proposition \ref{pr:eqneighbor} and Eq.\eqref{eq:dm},
we are motivated to recursively encode multiple r-digraphs with the shared 
edges in $\hat{\mathcal E}_{e_q}^\ell$ layer by layer.
The full process is in Algorithm~\ref{alg:gnn}.

\begin{algorithm}[ht]
	\caption{RED-GNN: recursive r-digraph encoding.}
	\label{alg:gnn}
	\small
	\begin{algorithmic}[1]
		\STATE initialize $\bm{h}_{e_q}^{0}\!(e_q, r_q)=\bm0$ and the entity set $\mathcal V_{e_q}^{0}=\{e_q\}$;
		\FOR{$\ell=1\dots L$}
		\STATE collect the $\ell$-hop edges $\hat{\mathcal E}_{e_q}^{\ell}\!=\!\{(e_s, r, e_o)\!\in\!\mathcal F| e_s\!\in\!\mathcal V_{e_q}^{\ell-1}\}$ \\
		and  entities $\mathcal V_{e_q}^{\ell} \!=\! \{e_o| (e_s, r, e_o)\!\in\! \hat{\mathcal E}_{e_q}^{\ell}\!\}$; \label{step:nodes}
		\STATE message passing for entities $e_o\in\mathcal V_{e_q}^\ell$: \\
		 $\bm{h}_{e_o}^{\ell}(e_q, r_q) = \delta
		\Big(\bm{W}^{\ell} \cdot \sum\nolimits_{(e_s, r, e_o)\in\hat{\mathcal E}_{e_q}^{\ell}} \phi\big(\bm h^{\ell\!-\!1}_{e_s}(e_q,r_q), \bm h_r^\ell\big)\Big)$;
		\ENDFOR
		\STATE assign $\bm h_{e_a}^L\!(e_q,r_q)=\bm 0$ for all $e_a\notin \mathcal V_{e_q}^L$;
		\RETURN $\bm h_{e_a}^L\!(e_q, r_q)$ for all $e_a\in\mathcal V$. \label{step:return}
	\end{algorithmic}
\end{algorithm}

Initially,
only $e_q$ is visible in $\mathcal V_{e_q}^{0}$.
In the $\ell$-th layer,
we collect the edges $\hat{\mathcal E}_{e_q}^{\ell}$ and entities $\mathcal V_{e_q}^{\ell}$
in step~\ref{step:nodes}.
Then, 
the message passing is constrained 
to obtain the representations for $e_o\!\in\!\mathcal V_{e_q}^\ell$
through the edges in ${\mathcal E}_{e_q}^{\ell}$.
Finally,
the
representations $\bm h_{e_a}^L\!(e_q,r_q)$,
encoding $\mathcal G_{e_q, e_a|L}$
for all the entities $e_a\in\mathcal V$, 
are returned in step~\ref{step:return}.
The 
recursive encoding can be more efficient with 
shared edges in $\hat{\mathcal E}_{e_q}^{\ell}$ and fewer loops.
It learns the same representations as Algorithm~\ref{alg:simple} as 
guaranteed by Proposition~\ref{pr:identical}.

\begin{prop}
	\label{pr:identical}
	Given
	the same triple $(e_q, r_q, e_a)$,
	the structures encoded in $\bm h_{e_a}^L\!(e_q, r_q)$ by Algorithm~\ref{alg:simple} and Algorithm~\ref{alg:gnn} are identical.
\end{prop}

\subsection{Interpretable reasoning with r-digraph}
\label{ssec:gnn}

As the construction of $\mathcal G_{e_q, e_a|L}$ is independent with
the query relation $r_q$,
how to encode $r_q$ is another problem to address.
Given different triples sharing the same r-digraph,
e.g.
\textit{(Sam, starred, Spider-2)}
and
\textit{(Sam, directed, Spider-2)},
the local evidence we use for reasoning is different.
To capture the query-dependent knowledge from the r-digraphs
and discover interpretable local evidence,
we use the attention mechanism \cite{velivckovic2017graph}
and encode $r_q$ into the attention weight
to control the importance of different edges in $\mathcal G_{e_q, e_a|L}$.
The message passing function is specified as
\begin{equation}
\!\!\!{\bm{h}}_{e_o}^{\ell}\!(e_q, \!r_q)
\!=\! \delta
\Big(\bm W^\ell \!\cdot \!\sum\nolimits_{(e_s, r, e_o)\in{\hat{\mathcal E}}_{e_q}^{\ell}} \!\!\!\alpha_{e_s,r,e_o\!|r_q}^{\ell}\!
\big(\bm{h}_{e_s}^{\ell-1}(e_q, \! r_q) + \bm{h}_r^{\ell} \big) \!
\Big),
\label{eq:aggregator}
\end{equation}
where the attention weight $\alpha_{e_s, r,e_o|r_q}^{\ell}$ on edge $(e_s, r, e_o)$ is
\begin{equation}
\vspace{-4px}
\!\!\!\alpha_{e_s,r,e_o\!|r_q}^{\ell} \!=\! 
\sigma
\left( 
(\bm w_{\alpha}^{\ell})^\top
\text{ReLU}
\left( 
\bm W_{\alpha}^{\ell} \!\cdot\!
\big(
\bm{h}_{e_s}^{\ell-1}(e_q, r_q)
\!\oplus\!
\bm{h}_r^{\ell}
\!\oplus\! 
\bm{h}_{r_q}^{\ell}
\big)
\right) 
\right) ,
\label{eq:attention}
\end{equation}
with $\bm w_{\alpha}^{\ell}\!\in\!\mathbb R^{d_\alpha}$, $\bm W_{\alpha}^{\ell}\!\in\!\mathbb R^{d_\alpha\times 3d}$, and
$\oplus$ is the concatenation operator.
Sigmoid function $\sigma$ is used rather than softmax attention~\cite{velivckovic2017graph}
to ensure multiple edges can be selected in the same neighborhood.

After $L$ layers' aggregation by \eqref{eq:aggregator},  
the representations $\bm{h}_{e_a}^{L}\!(e_q, r_q)$ can encode essential
information for scoring $(e_q, r_q, e_a)$.
Hence,
we design a simple scoring function 
\begin{equation}
f(e_q, r_q, e_a) = \bm{w}^{\top} \bm{h}_{e_a}^{L}\!(e_q, r_q),
\end{equation}
with $\bm{w} \in\mathbb R^{d}$.
We associate the multi-class log-loss~\cite{lacroix2018canonical} with each training triple $(e_q, r_q, e_a)$,
i.e., 
\begin{align}
\!\!\!\!\!\sum\nolimits_{(e_q, r_q, e_a) \in\mathcal{T}_{\text{tra}}}
\!\Big(\! 
- f(e_q, r_q, e_a) +\log \big(\!\sum\nolimits_{\forall e\in\mathcal V} e^{f(e_q, r_q, e)} \big)
\Big) .
\label{eq:loss}
\end{align}
The first part 
in  \eqref{eq:loss}
is the score of the positive triple $(e_q, r_q, e_a)$ in $\mathcal{T}_{\text{tra}}$, the set of training queries,
and the second part contains the scores of all triples with the same query $(e_q, r_q, ?)$.
The model parameters 
$\bm\Theta =$ 
$\big\{ \{ \bm W^{\ell} \}$, 
$\{ \bm w_{\alpha}^{\ell} \}$, 
$\{ \bm W_{\alpha}^{\ell} \}$, 
$\{  \bm{h}_r^{\ell} \}$, 
$\bm w \big \}$
are randomly initialized and
are optimized
by minimizing \eqref{eq:loss}
with stochastic gradient descent~\cite{kingma2014adam}.

We provide Theorem~\ref{theo:interp}
to show that if a set of relational paths  
are strongly correlated with the query triple,
they can be identified by the attention weights in RED-GNN,
thus being interpretable.
This theorem is also empirically illustrated in
Section~\ref{sec:exp:visual}.

\begin{theorem}
	\label{theo:interp}
	Given a triple $(e_q, r_q, e_a)$,
	let $\mathcal P$ be a set of relational paths
	$e_q\!\rightarrow_{r^1_i}\!\cdot\!\rightarrow_{r^2_i}\!\cdots\!\rightarrow_{r^{L}_i}\!e_a$,
	that are generated by a set of rules between $e_q$ and $e_a$ with the form
	\begin{equation*}
		r_i^1(X, Z_1)\wedge r_i^2(Z_1, Z_2)\wedge \cdots \wedge r_i^L(Z_{L-1},Y)\rightarrow r_q (X,Y),
	\end{equation*}
	where $X, Y, Z_1, \dots, Z_{L-1}$ are variables that are bounded by unique entities.
	Denote $\mathcal G_{\mathcal P}$ as the r-digraph constructed by $\mathcal P$.
	There exists a parameter setting $\bm \Theta$ and a threshold $\theta\in(0,1)$ for RED-GNN that
	$\mathcal G_{\mathcal P}$ can equals to
	the 
	r-digraph, whose edges have attention weights $\alpha_{e_s,r,e_o|r_q}^{\ell} > \theta$,
	in $\mathcal G_{e_q, e_a|L}$.
\end{theorem}

\subsection{Inference complexity}
\label{ssec:complexity}

In this part,
we compare the inference complexity of different GNN-based methods.
When reasoning on the query $(e_q, r_q, ?)$,
we need to evaluate $|\mathcal V|$ triples with the different answer entity $e\in\mathcal V$.
We assume the average degree as $\bar{D}$ and average number of edges in each layer of r-digraph as $\bar{E}$.
\begin{itemize}[leftmargin=*]
	\item RED-Simp (Algorithm~\ref{alg:simple}).
	For construction,
	the main cost,
	which is $O\big(\min(\bar{D}^L, |\mathcal F|L)\big)$
	with the worst case cost $O(|\mathcal F|L)$,
	comes from extracting the neighborhoods
	of $e_q$ and $e_a$.
	Along with encoding,
	the total cost is
	$O\big(|\mathcal V|\cdot(\min(\bar{D}^L, |\mathcal F|L)+ d\bar{E}L)\big)$.
	
	\item RED-GNN (Algorithm~\ref{alg:gnn}).
	Since the full computation is conducted on
	$\hat{\mathcal E}_{e_q}^\ell$ and $\mathcal V_{e_q}^\ell$,
	thus the cost is the number of edges times the dimension $d$,
	i.e. $O(d\cdot\min(\bar{D}^L, |\mathcal F|L))$ and $d\ll |\mathcal V|$.
	
	\item CompGCN.
	All the representations are aggregated in one pass with cost $O(d|\mathcal F|L)$.
	Then the scores for all the entities are computed with cost $O(|\mathcal V|d)$.
	The total cost is $O(d|\mathcal F|L+d|\mathcal V|)$.
	
	\item DPMPN.
	Denote the dimension and layer in the non-attentive GNN as $d_1, L_1$,
	the average sampled edges in the pruning procedure as $\bar{D}$.
	The main cost comes from the non-attentive GNN with cost $O(d_1|\mathcal F|L_1)$.
	The cost on sampled subgraph is $O(d\bar{D}^L)$.
	Thus, the total computation cost is $O(d_1|\mathcal F|L_1+d\bar{D}^L)$.
	
	\item GraIL.
	Denote $\bar{V}$ as the average number of entities in the subgraphs.
	The complexity in constructing the enclosing subgraph is $O(\bar{E}\log \bar{V})$ with $\bar{E}$ edges and $\bar{V}$ entities.
	The cost of the GNN module is $O(d\bar{E}L)$.
	Then
	the overall cost is $O\big(|\mathcal V|(\bar{E}\log{\bar{V}} + d\bar{E}L) \big)$.
\end{itemize}

In comparison,
we have RED-GNN $\approx$ CompGCN $<$ DPMPN $<$ RED-Simp $<$ GraIL
in terms of inference complexity.
The empirical evaluation is provided in Section~\ref{ssec:time}.

\section{Experiments}
\label{sec:exp}

All the experiments are written in Python with PyTorch framework~\cite{paszke2017automatic} 
and run on an RTX 2080Ti GPU with 11GB memory.

\subsection{Inductive reasoning}
\label{ssec:inductive}

Inductive reasoning is a hot 
research topic~\cite{hamilton2017inductive,sadeghian2019drum,teru2019inductive,yang2017differentiable}
as there are emerging new entities in the real-world applications,
such as new users, new items 
and new concepts \cite{zhang2019inductive}.
Being able to reason on unseen entities
requires the model to 
capture the semantic and local evidence
ignoring the identity of entities.

\begin{table*}[ht]
	\centering
	\caption{Inductive reasoning.
		Best performance is indicated by the bold face numbers.}
	\label{tab:induc}
	\vspace{-10px}
	\begin{tabular}{cc|cccc|cccc|cccc}
		\toprule
		&       & \multicolumn{4}{c|}{WN18RR} & \multicolumn{4}{c|}{FB15k-237}  & \multicolumn{4}{c}{NELL-995} \\
		&          & V1    & V2   & V3   & V4   & V1    & V2    & V3    & V4    & V1  & V2  & V3  & V4  \\   
		\midrule
		\multirow{5}{*}{MRR}   
		& RuleN   &  .668   & .645  & .368  & .624  &  .363  & .433  &  .439  & .429  & .615 &.385& .381 &.333\\
		&  Neural LP &	.649	&	.635	&	.361	&	.628	&	.325	&	.389	&	.400	&	.396	&	.610	&.361 & .367  &	.261	\\
		& DRUM      &  .666  &  .646    &   .380   &   .627   &   .333    &  .395     &   .402    &   .410    & .628  &  .365  & .375  &  .273  \\ \cmidrule{2-14}
		& GraIL     &     .627    &   .625   &   .323   &  .553    &  .279     &   .276    &    .251   &    .227  &  .481 & .297 &  .322  & .262  \\
		& \textbf{RED-GNN}      &   \textbf{.701}  &  \textbf{.690}   &  \textbf{.427}    &   \textbf{.651}   &     \textbf{.369}     &  \textbf{.469}     &  \textbf{.445}     &  \textbf{.442}  &  \textbf{.637}   & \textbf{.419} & \textbf{.436}  &  \textbf{.363}   \\ 
		\midrule
		\multirow{5}{*}{Hit@1 (\%)}
		& RuleN   &  63.5   &   61.1  &  34.7  &  59.2  &  30.9  & 34.7 & 34.5 & 33.8  & \textbf{54.5}&30.4& 30.3  &24.8\\
		&  Neural LP &	59.2	&	57.5	&	30.4	&	58.3	&	24.3	&	28.6	&	30.9	&	28.9	&	50.0	&		24.9   &	26.7 	& 13.7	\\
		& DRUM      &  61.3  &   59.5   &    33.0  &  58.6    &  24.7     &   28.4    &   30.8    &   30.9    & 50.0  &  27.1  &  26.2  &   16.3  \\ \cmidrule{2-14}
		& GraIL     &   55.4    &    54.2  &   27.8   &    44.3  &   20.5    &   20.2    &   16.5    &    14.3   &  42.5  & 19.9  &   22.4  &  15.3  \\ 
		& \textbf{RED-GNN}      &  \textbf{65.3}     &   \textbf{63.3}   &  \textbf{36.8}    &   \textbf{60.6}   &  \textbf{30.2}     &   \textbf{38.1}    &   \textbf{35.1}    &  \textbf{34.0}   & {52.5}   &  \textbf{31.9}  &  \textbf{34.5}  &  \textbf{25.9}  \\ 
		\midrule
		\multirow{5}{*}{Hit@10 (\%)}    
		& RuleN   &  73.0   &  69.4  &  40.7   &  68.1  & 44.6  & 59.9 & 60.0  &  60.5  & 76.0&51.4&53.1&48.4  \\
		&  Neural LP &	77.2	&	74.9 &	47.6	&	70.6	&	46.8	&	58.6	&	57.1	&	59.3	&	\textbf{87.1}	&		56.4   &	{57.6}	&53.9 \\
		& DRUM      &  77.7  &  74.7    &   47.7   &   70.2   &   {47.4}    &  59.5     &   {57.1}    &  59.3 &  87.3  &  {54.0}  & 57.7  &  {53.1}  \\  \cmidrule{2-14}
		& GraIL     &   76.0    &  77.6    &   40.9   &   68.7   &  42.9     &    42.4   &   42.4    &   38.9     &  56.5 & 49.6 & 51.8  &  50.6  \\
		& \textbf{RED-GNN}      &   \textbf{79.9}    &  \textbf{78.0}    &   \textbf{52.4}   &  \textbf{72.1}    &   \textbf{48.3}    &   \textbf{62.9}    &  \textbf{60.3}     &  \textbf{62.1}   &  {86.6} &  \textbf{60.1}  &  \textbf{59.4}  &  \textbf{55.6} \\ 
		\bottomrule
	\end{tabular}
\end{table*}

\begin{table*}[t]
	\centering
	\caption{Transductive reasoning. Best performance is indicated by the bold face numbers. `-' means unavailable results and results for methods with `*' are copied from the original papers.}
	\setlength\tabcolsep{2pt}
	\label{tab:transd}
	\vspace{-10px}
	\begin{tabular}{c|c|ccc|ccc|ccc|ccc|ccc}
		\toprule
		\multirow{2}{*}{type}  &   \multirow{2}{*}{models}   & \multicolumn{3}{c|}{Family} & \multicolumn{3}{c|}{UMLS}   &  \multicolumn{3}{c|}{WN18RR}   &  \multicolumn{3}{c|}{FB15k-237}   &  \multicolumn{3}{c}{NELL-995}  \\
		&  &  MRR     & Hit@1     & Hit@10   &	MRR     & Hit@1     & Hit@10   &MRR     & Hit@1     & Hit@10   &   MRR     &   Hit@1    &     Hit@10       &      MRR       & 		Hit@1      &      Hit@10     \\   \midrule
		\multirow{3}{*}{triple} & ConvE*   &  -  & - &  - & .94 & 92. &  96. &  .43  & 39. & 49.  & .325 & 23.7 & 50.1 & - & - & - \\
		&  RotatE  		&.921&86.6&98.8& .925&86.3&99.3& .477& 42.8&57.1	&	.337&24.1&53.3	& .508&44.8&60.8  \\ 
		&  QuatE   	&.941&89.6&99.1&  .944&90.5&99.3	& .480&44.0&55.1&	.350&25.6&53.8	&	.533&46.6&64.3  \\ 
		\midrule
		\multirow{4}{*}{path}    &  MINERVA   &  .885  &  82.5  & 96.1  &  .825&72.8&96.8&  .448&41.3&51.3	&	.293&21.7&45.6& .513&41.3&63.7\\ 
		&  Neural LP  & .924&87.1&{99.4}   &.745&62.7& 91.8  &.435&37.1&56.6	&	.252&18.9&37.5&\multicolumn{3}{c}{out ~ of ~memory}\\ 
		&  DRUM   & .934    & 88.1    & {99.6}   & .813   & 67.4   &  {97.6}&.486&42.5&58.6	&	.343&25.5&51.6	&\multicolumn{3}{c}{out ~ of ~memory}\\ 
		&  RNNLogic* & --  &  --  & -- & .842&77.2&96.5&	.483& 44.6&55.8	&	.344&25.2&53.0	&	--  &  --  & -- \\ 
		\midrule
		\multirow{4}{*}{GNN}  & pLogicNet*  &   --  &  --  & -- & .842&77.2&96.5&	.441 & 39.8 & 53.7	&	.332  &  23.7   &   52.8	&	--  &  --  & -- \\ 
		&  CompGCN  &  .933&88.3&{99.1}  & .927&86.7&\textbf{99.4} &  .479&44.3&54.6	&	.355&26.4&53.5	&	\multicolumn{3}{c}{out ~ of ~memory}	\\ 
		&  DPMPN   & .981&97.4&98.1  &.930&89.9& 98.2 &	.482&44.4&55.8	&	{.369}&\textbf{28.6}&53.0	&	.513&45.2&61.5  \\ 
		\cmidrule{2-17}
		&  \textbf{RED-GNN}   &  \textbf{.992}&\textbf{98.8}&\textbf{99.7} &\textbf{.964}&\textbf{94.6}&99.0	&  \textbf{.533}&\textbf{48.5}&\textbf{62.4}	& \textbf{.374}&28.3&\textbf{55.8}  &  \textbf{.543}&\textbf{47.6}&\textbf{65.1}  \\ 
		\bottomrule
	\end{tabular}
\end{table*}

\noindent 
\textbf{Setup.}
We follow the general inductive setting~\cite{sadeghian2019drum,teru2019inductive,yang2017differentiable}
where 
there are new entities in testing
and the relations are the same as those in training.
Specifically,
the training and testing contain two KGs 
$\mathcal K_{\text{tra}}=\{\mathcal V_{\text{tra}}, \mathcal R, \mathcal F_{\text{tra}}\}$
and $\mathcal K_{\text{tst}}=\{\mathcal V_{\text{tst}}, \mathcal R, \mathcal F_{\text{tst}}\}$,
with the same set of relations but disjoint sets of entities.
Three sets of triples $\mathcal T_{\text{tra}}/\mathcal T_{\text{val}}/\mathcal T_{\text{tst}}$, 
augmented with reverse relations,
are provided.
$\mathcal F_{\text{tra}}$ is used to predict $\mathcal T_{\text{tra}}$ and $\mathcal T_{\text{val}}$ for training and validation, respectively.
In testing,
$\mathcal F_{\text{tst}}$ is used to predict $\mathcal T_{\text{tst}}$.
Same as~\cite{sadeghian2019drum,teru2019inductive,yang2017differentiable},
we use the filtered ranking metrics,
i.e., mean reciprocal rank (MRR), Hit@1 and Hit@10
\cite{bordes2013translating},
to indicate better performance with larger values.

\noindent
\textbf{Baselines.}
Since training and testing contain disjoint sets of entities,
all the methods requiring the entity embeddings \cite{bordes2013translating,harsha2020probabilistic,sun2019rotate,vashishth2019composition,xu2019dynamically,zhang2019quaternion}
cannot be applied here.
We mainly compare with four methods:
1) RuleN~\cite{meilicke2018fine},
the discrete rule induction method;
2) Neural-LP~\cite{yang2017differentiable},
the first differentiable method for rule learning;
3) DRUM~\cite{sadeghian2019drum},
an improved work of Neural-LP~\cite{yang2017differentiable};
and 
4) GraIL~\cite{teru2019inductive},
which designs the enclosing subgraph for inductive reasoning.
MINERVA~\cite{das2017go},
PathCon~\cite{wang2021relational} and RNNLogic~\cite{qu2021rnnlogic}
can potentially work on this setting
but there lacks the customized source code for the inductive setting, thus not compared.

\noindent
\textbf{Hyper-parameters.}
For RED-GNN,
we tune the learning rate in $[10^{-4},$$ 10^{-2}]$,
weight decay in $[10^{-5},$$ 10^{-2}]$,
dropout rate in $[0,$$ 0.3]$,
batch size in $\{5,10,$$ 20,$$ 50,$$ 100\}$,
dimension $d$ in $\{32,$$ 48,$$ 64,$$ 96\}$,
$d_\alpha$ for attention in $\{3,$$5\}$,
layer $L$ in $\{3,$$ 4,$$ 5\}$,
and activation function $\delta$ in \{identity, tanh, ReLU\}.
Adam~\cite{kingma2014adam} is used as the optimizer.
The best hyper-parameter settings are selected by the MRR metric
on $\mathcal T_{\text{val}}$ with maximum training epochs of 50.
For RuleN,
we use their implementation
with default setting.
For Neural LP
and DRUM,
we tune the learning rate in $[10^{-4},$$ 10^{-2}]$,
dropout rate in $[0,$$ 0.3]$,
batch size in $\{20,$$ 50,$$ 100\}$,
dimension $d$ in $\{64,$$ 128\}$,
layer $L$ of RNN in $\{1,2\}$,
and number of steps in $\{2,3,4,5\}$.
For GraIL
we tune the learning rate in $[10^{-5},$$ 10^{-2}]$,
weight decay in $[10^{-6},$$ 10^{-3}]$,
batch size in $\{8,16,32\}$,
dropout rate in $[0,$$ 0.4]$,
edge\_dropout rate in $[0, 0.6]$,
GNN aggregator among \{sum, MLP, GRU\}
and hop numbers among $\{2,3,4\}$.
The training epochs are all set as 50.

\noindent
\textbf{Tie policy.}
In evaluation,
the tie policy is important.
Specifically,
when there are triples with the same rank,
choosing the largest rank and smallest rank in a tie will lead to rather different results~\cite{sun2020re}.
Considering that we give the same score $0$
for triples where $\mathcal G_{e_q, e_a|L}=\emptyset$,
there will be a concern of the tie policy. 
Hence,
we use the average rank among the triples in tie as suggested \cite{rossi2020knowledge}.

\noindent
\textbf{Datasets.}
We use the 
benchmark dataset in~\cite{teru2019inductive},
created on 
{WN18RR}~\cite{dettmers2017convolutional}, 
{FB15k237}~\cite{toutanova2015observed} 
and NELL-995~\cite{xiong2017deeppath}.
Each dataset includes four versions with different groups of triples.
Please refer to \cite{teru2019inductive} for more details.

\noindent
\textbf{Results.}
The performance is shown in Table~\ref{tab:induc}.
First,
GraIL is the worst among all the methods
since the enclosing subgraphs do not learn well of the relational structures
that can be generalized to unseen entities
(more details in Appendix~\ref{app:grail}).
Second,
there is not absolute winner among the rule-based methods
as different rules adapt differently to these datasets.
In comparison,
RED-GNN outperforms the baselines
across all the benchmarks.
Based on Thereom~\ref{theo:interp},
the attention weights can help to adaptively learn
correlated relational paths for different datasets,
and preserve the structural patterns at the same time.
In some cases,
the Hit@10 of RED-GNN is slightly worse than the rule-based methods
since it may overfit to the top-ranked samples.

\subsection{Transductive reasoning}
\label{ssec:transductive}

Transductive reasoning,
also known as KG completion~\cite{trouillon2017knowledge,wang2017knowledge}, 
is another general setting in the literature.
It evaluates the models' ability to learn the patterns on an incomplete KG.

\noindent
\textbf{Setup.}
In this setting,
a KG $\mathcal K=\{\mathcal V, \mathcal R, \mathcal F\}$ 
and the query triples $\mathcal T_{val}/\mathcal T_{\text{tst}}$,
augmented with reverse relations, are given.
For the triple-based method,
triples in $\mathcal F$ are used for training,
and $\mathcal T_{val}/\mathcal T_{\text{tst}}$ are used for inference.
For the others,
$\nicefrac{3}{4}$ of the triples in $\mathcal F$ are used to extract paths/subgraphs 
to predict the remaining $\nicefrac{1}{4}$ triples in training,
and the full set $\mathcal F$ is then used to predict $\mathcal T_{val}/\mathcal T_{\text{tst}}$ in inference 
\cite{yang2017differentiable,sadeghian2019drum}.
We use the same filtered ranking metrics 
with the same 
tie policy 
in Section~\ref{ssec:inductive},
namely MRR, Hit@1 and Hit@10.

\noindent
\textbf{Baselines.}
We compare 
RED-GNN
with
the triple-based methods
ConvE~\cite{dettmers2017convolutional},
RotatE~\cite{sun2019rotate} and
QuatE~\cite{zhang2019quaternion};
the path-based methods 
MINERVA~\cite{das2017go},
Neural LP~\cite{yang2017differentiable},
DRUM~\cite{sadeghian2019drum}
and RNNLogic~\cite{qu2021rnnlogic};
MLN-based method pLogicNet~\cite{qu2019probabilistic}
and the GNN-based methods
CompGCN~\cite{vashishth2019composition}
and DPMPN~\cite{xu2019dynamically}.
RuleN~\cite{meilicke2018fine}
is not compared here since it has been shown to be worse than DRUM~\cite{sadeghian2019drum}
and RNNLogic~\cite{qu2021rnnlogic} in this setting.
p-GAT is not compared as their results are evaluated on a problematic framework~\cite{sun2020re}.
GraIL~\cite{teru2019inductive} is not compared since
it is computationally intractable 
on large graphs
(see Section~\ref{ssec:time}).

\noindent
\textbf{Hyper-parameters.}
The tuning ranges of hyper-parameters of RED-GNN
are the same as those in the inductive reasoning.
For RotatE and QuatE,
we tune the dimensions in $\{100, 200, 500, 1000\}$,
batch size in $\{256, 512, 1024, 2048\}$,
weight decay in $[10^{-5},$$ 10^{-2}]$,
number of negative samples in
$\{64, 128, 256, 512, 1024\}$,
with training iterations of 100000.
For MINERVA, Neural LP, DRUM and DPMPN,
we use their default setting provided.
For CompGCN,
we choose the decoder as ConvE,
the  operator as  circular correlation
as suggested \cite{vashishth2019composition},
and tune
the learning rate in $[10^{-4},$$ 10^{-2}]$,
layer in $\{1,2,3\}$
and 
dropout rate in $[0,$$ 0.3]$
with training epochs of 300.

\noindent
\textbf{Datasets.}
Five
\footnote{
	Data from
	\url{https://github.com/alisadeghian/DRUM/tree/master/datasets} and \url{https://github.com/thunlp/OpenKE/tree/OpenKE-PyTorch/benchmarks/NELL-995}.}
datasets are used including
{Family}~\cite{kok2007statistical},  
{UMLS}~\cite{kok2007statistical}, 
{WN18RR}~\cite{dettmers2017convolutional}, 
{FB15k237}~\cite{toutanova2015observed}
and {NELL-995}~\cite{xiong2017deeppath}.
We provide the statistics of entities, relations and split of triples in Table~\ref{tab:trans-data}.

\begin{table}[ht]
	\centering
	\caption{Statistics of transductive reasoning datasets. Note that NELL-995* is different as the version in~\cite{das2017go}
		since the training triples contains valid and test triples there.}
	\label{tab:trans-data}
	\vspace{-10px}
	\begin{tabular}{c|ccccc}
		\toprule
		& $|\mathcal V|$   & $|\mathcal R|$ & $|\mathcal F|$   & $|\mathcal T_{\text{val}}|$  & $|\mathcal T_{\text{tst}}|$   \\
		\midrule
		Family    & 3,007  & 12   & 23,483  & 2,038  & 2,835  \\
		UMLS      & 135    & 46   & 5,327   & 569    & 633    \\
		WN18RR    & 40,943 & 11   & 86,835  & 3,034  & 3,134  \\
		FB15k-237 & 14,541 & 237  & 272,115 & 17,535 & 20,466 \\
		NELL-995* & 74,536 & 200 & 149,678 &  543  & 2,818 \\
		\bottomrule
	\end{tabular}
\end{table}

\noindent
\textbf{Results.}
As in Table~\ref{tab:transd},
the triple-based methods are better than the path-based ones
on Family and UMLS,
and is comparable with DRUM and RNNLogic on WN18RR, FB15k-237.
The entity embeddings can implicitly preserve local information around entities,
while the path-based methods may loss the structural patterns.
CompGCN performs similar as the triple-based methods
since
it mainly relies on the aggregated embeddings and the decoder scoring function.
Neural LP, DRUM and CompGCN run out of memory on NELL-995 with $74k$ entities
due to the use of full adjacency matrix.
For DPMPN, 
the entities in the pruned subgraph is more informative than that in CompGCN,
thus has better performance.
For RED-GNN,
it is better than all the baselines indicated by the MRR metric.
These demonstrate that the r-digraph can not only transfer well to unseen entities,
but also capture the important patterns in incomplete KGs
without using entity embeddings.

\subsection{Complexity analysis}
\label{ssec:time}

\begin{figure}[ht]
	\centering
	\vspace{-4px}
	\subfigure[Running time.]
	{\includegraphics[height=3.24cm]{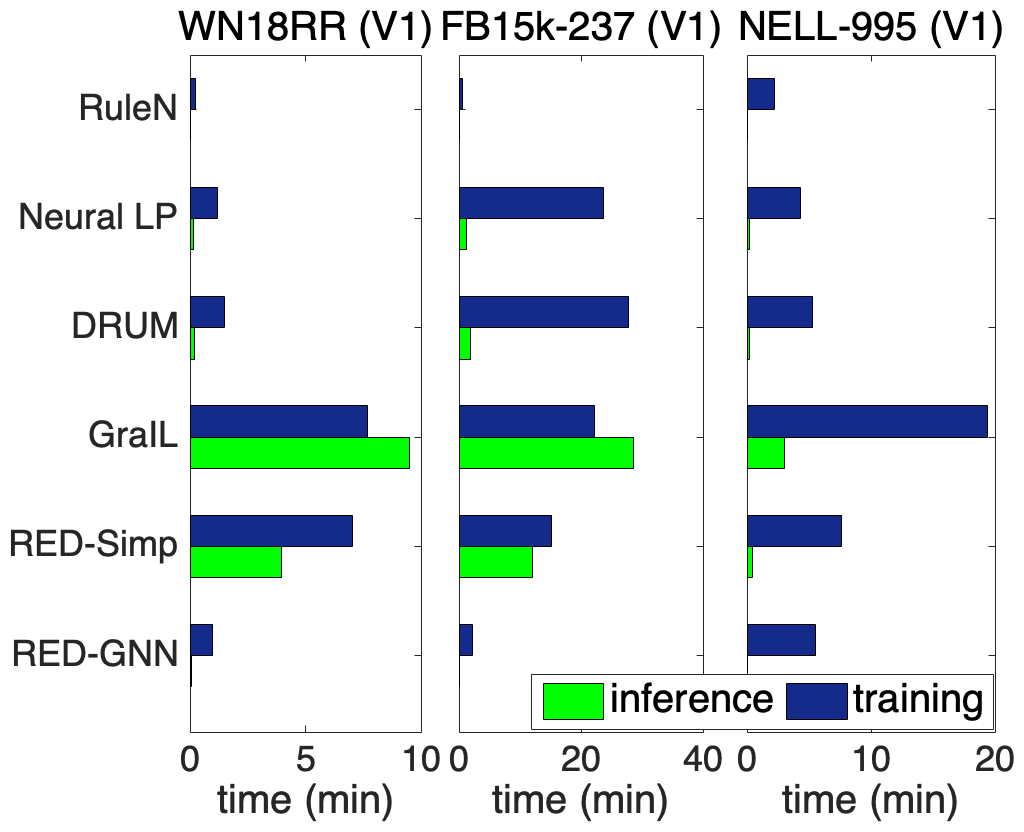} \hfill
	\includegraphics[height=3.24cm]{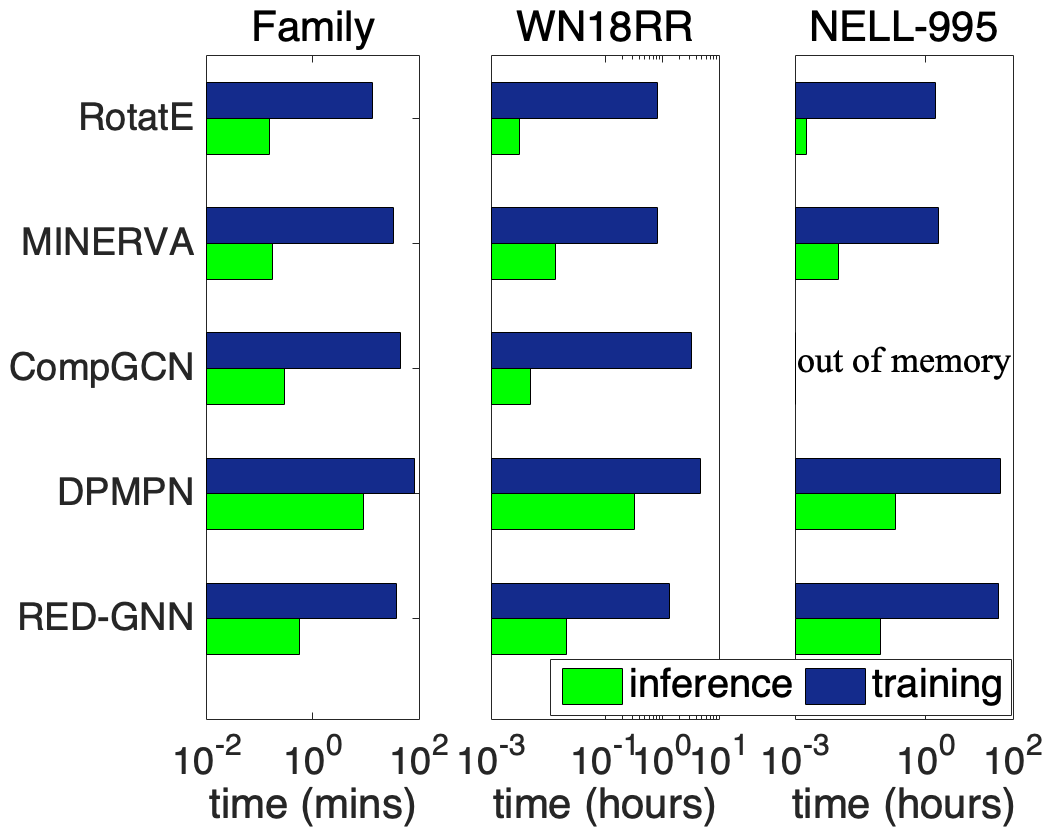}\label{fig:time-part}}
 	\vspace{-4px}
	
	\subfigure[Learning curve.]
	{\includegraphics[height=3.24cm]{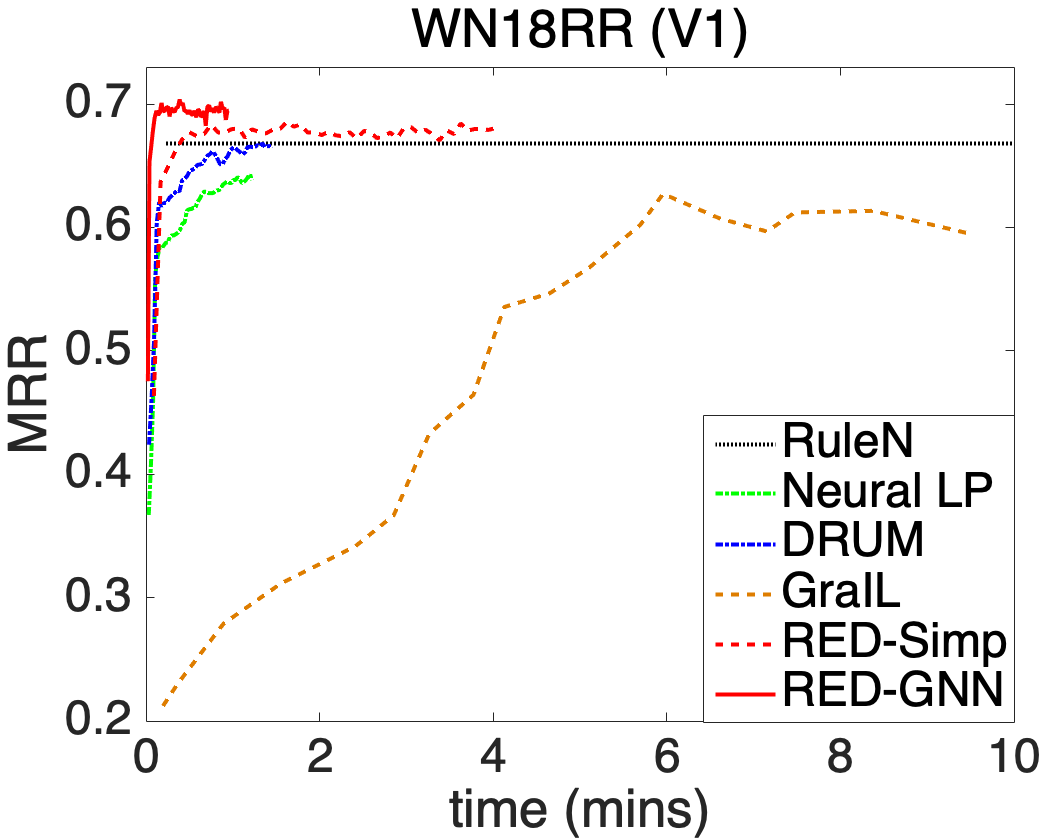}  \hfill
	\includegraphics[height=3.24cm]{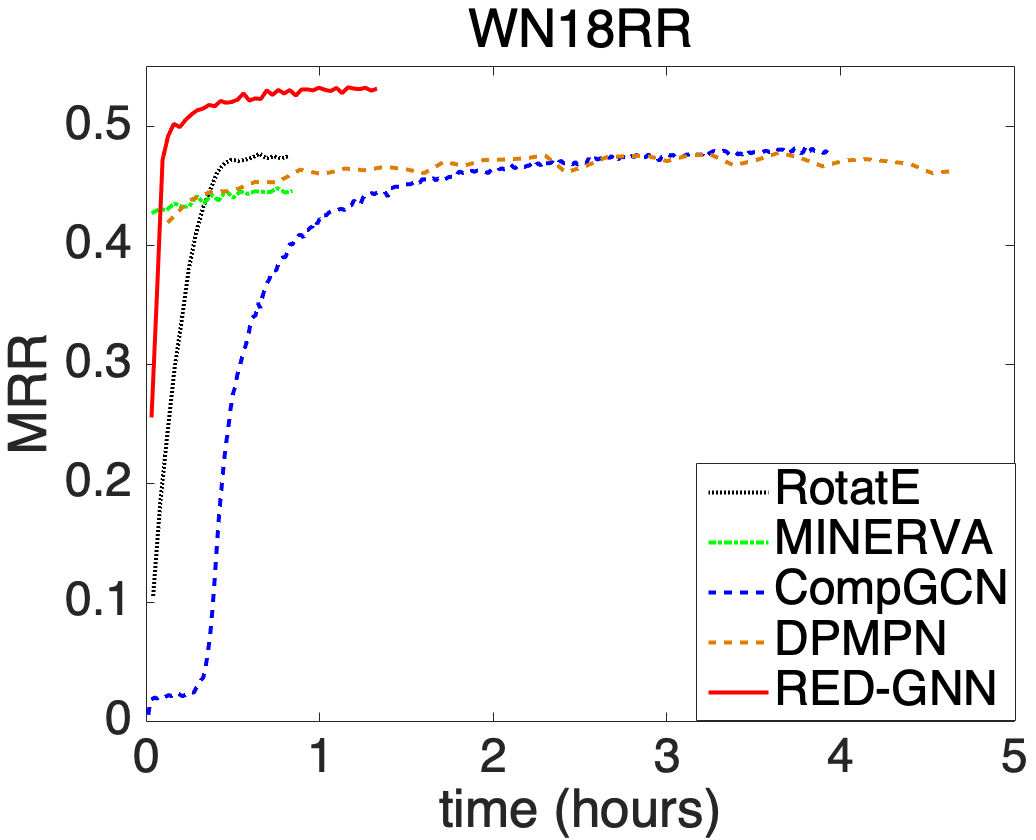}\label{fig:time-curve}}
 	\vspace{-4px}

	\subfigure[Model parameters.]
	{\includegraphics[height=3.24cm]{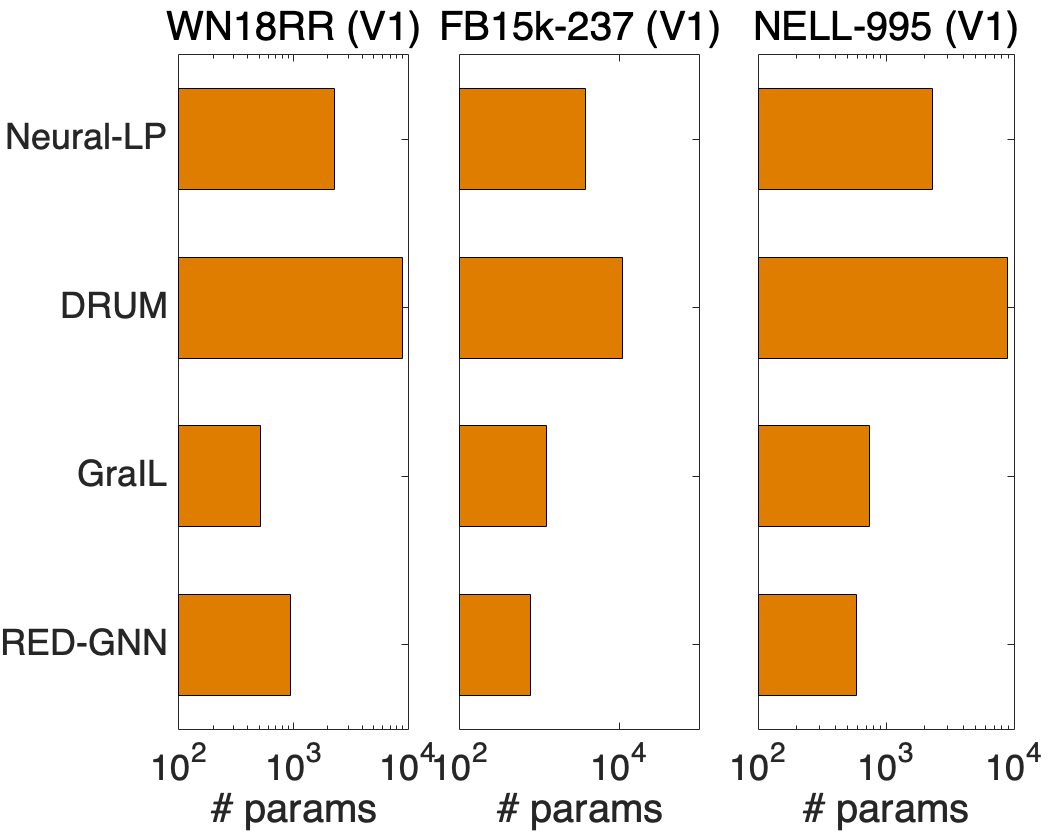} \hfill
		\includegraphics[height=3.24cm]{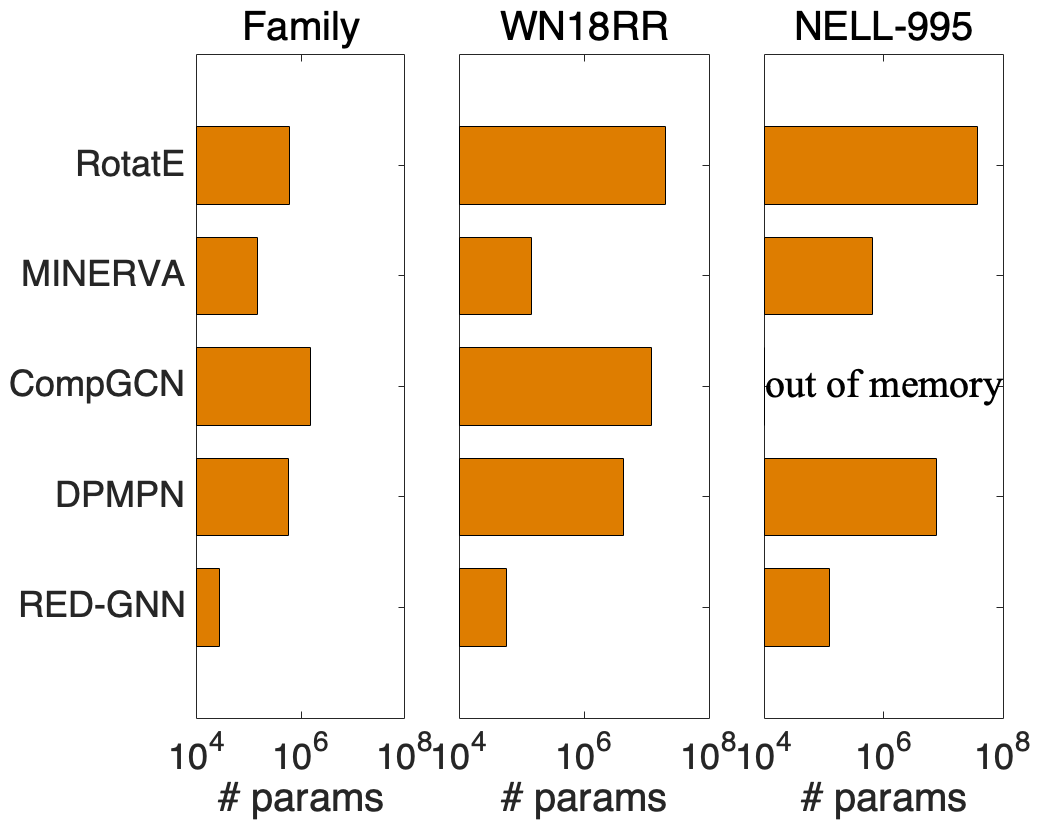}\label{fig:params}}

	\vspace{-13px}
	\caption{Running time analysis and learning curve.
		Left: inductive setting;
		Right: transductive setting.}
	\label{fig:running-time}
\end{figure}

We compare the complexity
in terms of 
running time and parameter size 
of different methods in this part.
We show the training time and the inference time on $\mathcal T_{\text{tst}}$ for each method
in Figure~\ref{fig:time-part},
learning curves in Figure~\ref{fig:time-curve},
and model parameters in Figure~\ref{fig:params}.

\noindent
\textbf{Inductive reasoning.}
We compare RuleN, Neural LP, DRUM, GraIL
and RED-GNN
on WN18RR (V1), FB15k-237 (V1) and NELL-995 (V1).
Both training and inference are very efficient in RuleN.
Neural LP and DRUM have similar cost but are more expensive than RED-GNN
by using the full adjacency matrix.
For GraIL,
both training and inference are very expensive
since they require bidirectional sampling to extract the subgraph
and then compute for each triple.
As for model parameters,
RED-GNN and GraIL have similar amount of parameters,
less than Neural-LP and DRUM.
Overall,
RED-GNN is more efficient than the differentiable methods Neural LP, DRUM and GraIL.

\noindent
\textbf{Transductive reasoning.}
We compare RotatE, MINERVA, CompGCN, DPMPN and RED-GNN
on Family, WN18RR and NELL-995.
Due to the simple training framework on triples,
RotatE is the fastest method.
MINERVA is more efficient than GNN-based methods
since the sampled paths can be efficiently modeled in sequence.
CompGCN, with fewer layers,
has similar cost with RED-GNN as its computation is on the whole graph.
For DPMPN,
the pruning is expensive and it has two GNNs working together,
thus it is more expensive than RED-GNN.
GraIL is hundreds of times more expensive than RED-GNN,
thus intractable on the larger KGs in this setting.
Since RED-GNN does not learn entity embeddings,
it has much less parameters compared with the other four methods.

\subsection{Case study: learned r-digraphs}
\label{sec:exp:visual}

\begin{figure}[ht]
	\centering
	\subfigure[Family.]
	{\includegraphics[width=\columnwidth]{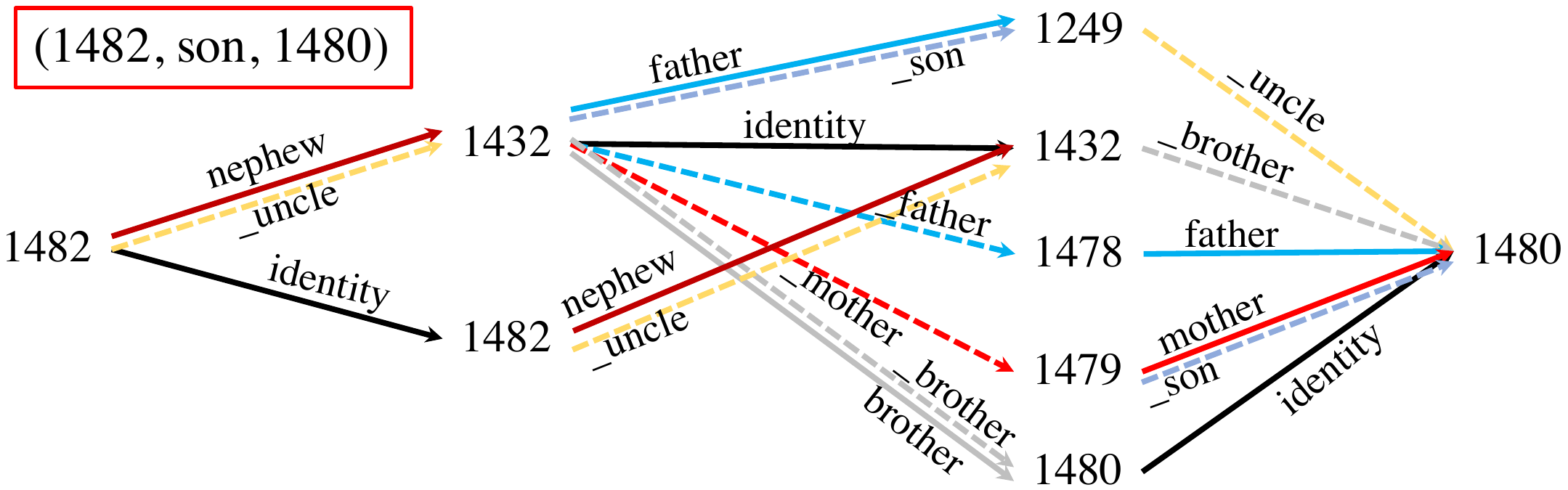}\label{fig:vis:family}}
	\vspace{-5px}
	
	\subfigure[UMLS.]{\includegraphics[width=1\columnwidth]{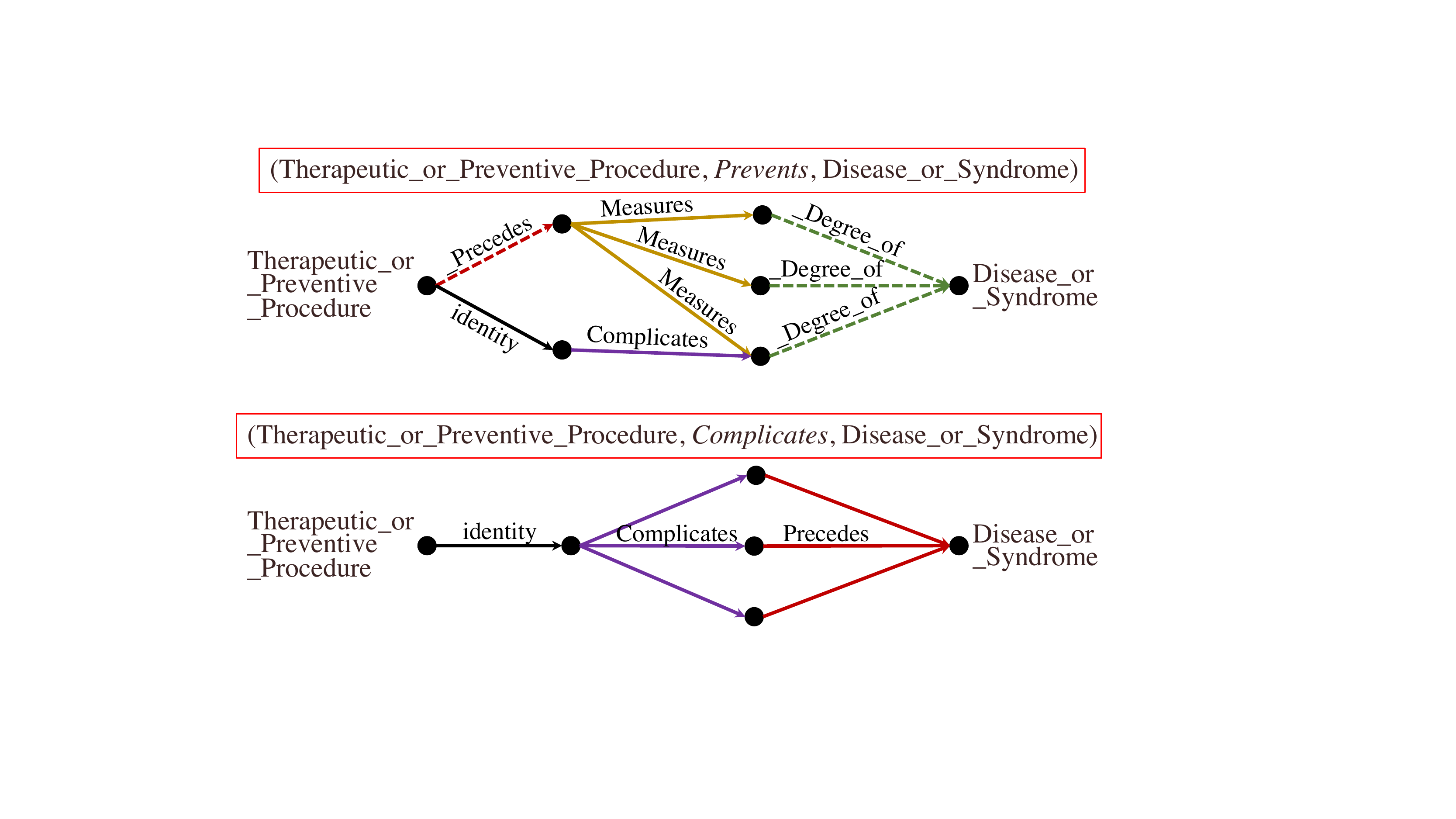}\label{fig:vis:umls2}}
	\vspace{-10px}
	\caption{Visualization of the learned structures. Dashed lines mean inverse relations. 
		The query triples are indicated by the red rectangles. 
		Due to space limitation, 
		entities in UMLS dataset are shown as black circles (best viewed in color).}
	\label{fig:visual}
\end{figure}

We visualize some exemplar learned r-digraphs by RED-GNN 
with $L\!=\!3$ on the Family and UMLS datasets.
Given the triple $(e_q, r_q, e_a)$,
we remove the edges in $\mathcal G_{e_q, e_a|L}$ whose attention weights are less than $0.5$,
and extract the remaining parts.
Figure~\ref{fig:vis:family} shows one triple that DRUM fails.
As shown, inferring id-$1482$ as the son of id-$1480$
requires the knowledge that id-$1480$ is the only brother of the uncle id-$1432$ from the local structure.
Figure~\ref{fig:vis:umls2} shows the example with 
the same $e_q$ and $e_a$
sharing the same digraph $\mathcal G_{e_q, e_a|L}$. 
As shown,
RED-GNN can learn distinctive structures for different query relations,
which caters to Theorem~\ref{theo:interp}.
The examples in Figure~\ref{fig:visual} demonstrate that RED-GNN is interpretable.
We provide more examples and the visualization algorithm in Appendix~\ref{app:visual}.

\subsection{Ablation study}

\noindent
\textbf{Variants of RED-GNN.}
Table~\ref{tab:abla} shows the performance of different variants.
First,
we study the impact of 
removing $r_q$ in attention (denoted as Attn-w.o.-$r_q$).
Specifically,
we remove 
$\bm{h}_{r_q}^{\ell}$ from the attention weight $\alpha_{e_s,r,e_o|r_q}^{\ell}$ in \eqref{eq:attention}
and change the scoring function to $\!\bm w^\top\!(\bm{h}_{e_a}^{L}\!(e_q, r_q)\!\oplus\! \bm{h}_{r_q}\!)$,
with $\bm w\!\in\!\!\mathbb R^{2d}\!$.
Since the attention $\alpha_{e_s,r,e_o|r_q}^{\ell}$
aims to
figure out the important edges in $\mathcal G_{e_q,e_a|L}$,
the learned structure will be less informative without the control of the query relation $r_q$,
thus has poor performance.

\begin{table}[ht]
	\centering
	\caption{Comparison of different variants of RED-GNN.}
	\label{tab:abla}
	\vspace{-10px}
	\setlength\tabcolsep{3.5pt}
	\begin{tabular}{c|cc|cc|cc}
		\toprule
		& \multicolumn{2}{c|}{WN18RR (V1)}  & \multicolumn{2}{c|}{FB15k-237 (V1)}  & \multicolumn{2}{c}{NELL (V1)}\\
		methods &  MRR &  H@10  &  MRR  & H@10  &  MRR   & H@10  \\\midrule
		Attn-w.o.-$r_q$ & .659 &   78.3 & .268   & 37.6 & .517 &  73.4 \\
		RED-Simp & .683  &    79.6 & .311 &   45.3  & .563   &  75.8  \\ 
		RED-GNN & {.701} &   79.9  & {.369}  &   48.3  & {.637}    &  86.6  \\
		\bottomrule
	\end{tabular}
\end{table}

Second,
we 
replace Algorithm~\ref{alg:gnn}
in RED-GNN with the simple solution in Algorithm~\ref{alg:simple},
i.e. RED-Simp.
Due to the efficiency issue, the loss function \eqref{eq:loss},
which requires to compute the scores over many negative triples,
cannot be used.
Hence, we use the margin ranking loss with one negative sample
as in GraIL~\cite{teru2019inductive}.
As in Table~\ref{tab:abla},
the performance of RED-Simp is weak than RED-GNN since
the multi-class log loss  
is better than loss functions with negative sampling~\cite{lacroix2018canonical,ruffinelli2020you,zhang2019nscaching}.
RED-Simp still outperforms GraIL in Table~\ref{tab:induc}
since the r-digraphs are better structures for reasoning.
The running time of RED-Simp is in 
Figure~\ref{fig:time-part} and \ref{fig:time-curve}.
Algorithm~\ref{alg:simple} is much more expensive than Algorithm~\ref{alg:gnn}
but is cheaper than GraIL
since GraIL needs
the Dijkstra algorithm to label the entities.

\begin{figure}[ht]
	\centering
	\includegraphics[height=3cm]{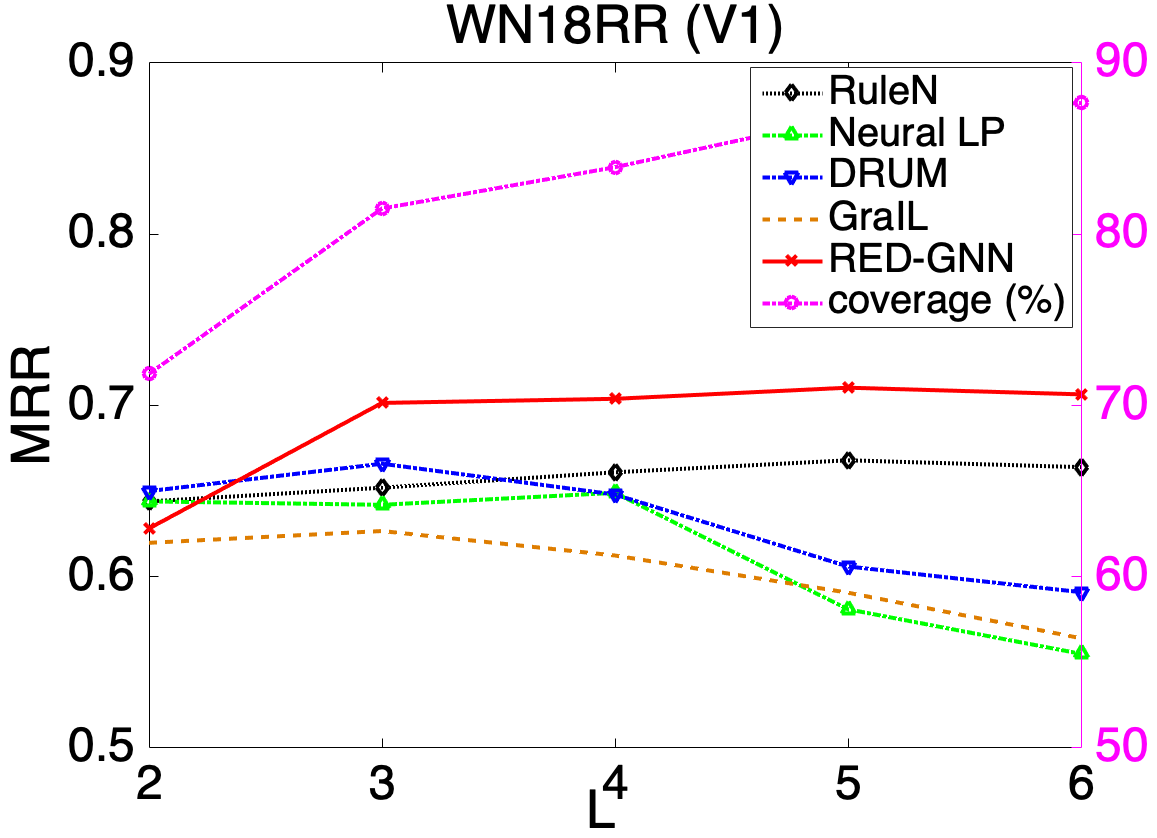}  \hfill
	\includegraphics[height=3cm]{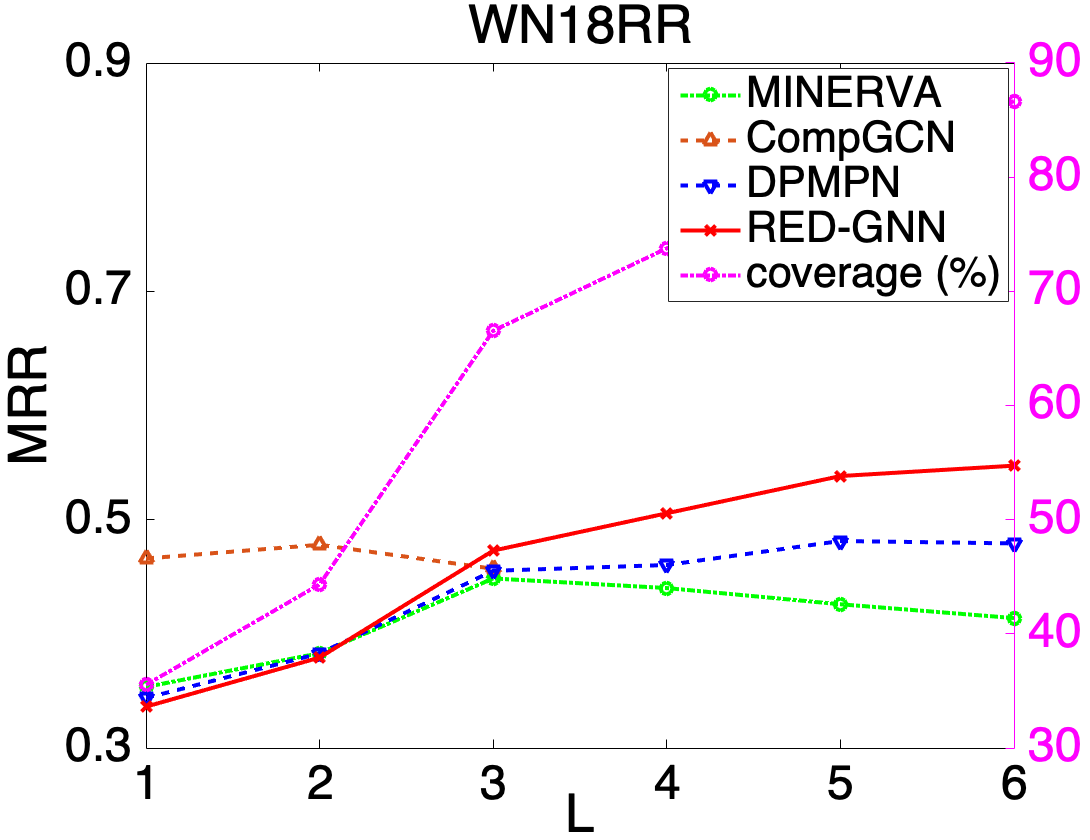}
	\vspace{-5px}
	\caption{The MRR performance with different $L$ and coverage of triples within $L$ steps.}
	\label{fig:depth}
	\vspace{-6px}
\end{figure}

\noindent
\textbf{Depth of models.}
In Figure~\ref{fig:depth},
we show the influence of testing MRR with different layers or steps $L$ in the left $y$-axis.
The coverage (in \%) of testing triples $(e_q, r_q, e_a)$
where $e_a$ is visible from $e_q$ in $L$ steps,
i.e., $e_a\in\mathcal V_{e_q}^L$,
is shown in the right $y$-axis. 
Intuitively,
when $L$ increases,
more triples will be covered,
paths or subgraphs between $e_q$ and $e_a$
then contain richer information,
but will be harder to learn.
As shown,
the performance of DRUM, Neural LP and MINERVA
decreases for $L\geq 4$.
CompGCN runs out of memory when $L>3$
and it is also hard to capture complex structures with $L\!=\!3$.
When $L$ is too small,
e.g., $L\leq2$,
RED-GNN has poor performance 
mainly dues to limited information encoded in 
such small r-digraphs.
RED-GNN achieves the best performance for $L\geq3$
where the r-digraphs can contain 
richer information
and 
the important information for reasoning
can be effectively learned by \eqref{eq:aggregator}.
Since the computation cost significantly increases with $L$,
we tune $L \in \{3,4,5\}$ to balance the efficiency and effectiveness in practice.

\begin{table}[ht]
	\centering
	\caption{The per-distance evaluation for MRR on WN18RR v.s. 
		the length of shortest path.}
	\label{tab:distance}
	\vspace{-8px}
	\begin{tabular}{c|cccccc}
		\toprule
		{distance} & 1  & 2 & 3 & 4 & 5  &  $>$5 \\ \midrule
		{ratios (\%)}  & 34.9 & 9.3 & 21.5 & 7.5  & 8.9  & 17.9 \\ \midrule
		CompGCN &  .993  & .327 & .337&.062&.061& .016 \\
		DPMPN & .982&.381&.333&.102&.057&.001\\
		RED-GNN & .993 & .563 & .536 & .186 & .089  & .005 \\
		\bottomrule
	\end{tabular}
\end{table}

\noindent
\textbf{Per-distance performance.}
Note that
given a r-digraph with $L$ layers,
the information between two nodes that are not reachable with $L$ hops
cannot be propagated by Algorithm~\ref{alg:gnn}.
This may raise a concern about
the predicting ability 
of RED-GNN,
especially for triples not reachable in $L$ steps.
We demonstrate this is not a problem here.
Specifically,
given a triple $(e_q, r_q, e_a)$,
we compute the shortest distance from $e_q$ to $e_a$.
Then, the MRR performance is grouped in different distances.
We compare CompGCN ($L\!=\!2$),
DPMPN ($L\!=\!5$) and RED-GNN ($L\!=\!5$) in
Table~\ref{tab:distance}.
All the models have worse performance on triples with larger distance
and cannot well model the triples that are far away.
RED-GNN has the best per-distance performance within distance $5$.


%
%

\section{Conclusion}
\label{sec:conclude}

In this paper,
we introduce a novel relational structure,
i.e., r-digraph,
as a generalized structure of relational paths
for KG reasoning.
Individually computing on each r-digraph is 
expensive for the reasoning task $(e_q, r_q, ?)$.
Hence,
inspired by 
solving overlapping sub-problems by dynamic programming,
we propose RED-GNN as a variant of GNN,
to efficiently construct
and effectively learn the r-digraph.
We show that RED-GNN achieves the state-of-the-art
performance in both KG with inductive and transductive 
KG reasoning benchmarks.
The training and inference of RED-GNN are very efficient
compared with the other GNN-based baselines.
Besides,
interpretable structures for reasoning can be learned by RED-GNN.

Lastly,
a concurrent work NBFNet
\cite{zhu2021neural} proposes to recursively encode all the paths
between the query entity to multiple answer entities,
which is similar to RED-GNN.
However,
NBFNet uses the full adjacency matrix for propagation,
which is very expensive and requires 128GB GPU memory.
In the future work,
we can leverage the 
pruning technique in \cite{xu2019dynamically}
or 
distributed programming in~\cite{cohen2019scalable}
to apply RED-GNN on KG with extremely large scale.


\bibliographystyle{ACM-Reference-Format}
\bibliography{acmart}


\begin{thebibliography}{57}


\ifx \showCODEN    \undefined \def \showCODEN     #1{\unskip}     \fi
\ifx \showDOI      \undefined \def \showDOI       #1{#1}\fi
\ifx \showISBNx    \undefined \def \showISBNx     #1{\unskip}     \fi
\ifx \showISBNxiii \undefined \def \showISBNxiii  #1{\unskip}     \fi
\ifx \showISSN     \undefined \def \showISSN      #1{\unskip}     \fi
\ifx \showLCCN     \undefined \def \showLCCN      #1{\unskip}     \fi
\ifx \shownote     \undefined \def \shownote      #1{#1}          \fi
\ifx \showarticletitle \undefined \def \showarticletitle #1{#1}   \fi
\ifx \showURL      \undefined \def \showURL       {\relax}        \fi
\providecommand\bibfield[2]{#2}
\providecommand\bibinfo[2]{#2}
\providecommand\natexlab[1]{#1}
\providecommand\showeprint[2][]{arXiv:#2}

\bibitem[\protect\citeauthoryear{Abujabal, Saha~Roy, Yahya, and
  Weikum}{Abujabal et~al\mbox{.}}{2018}]%
        {abujabal2018never}
\bibfield{author}{\bibinfo{person}{A. Abujabal}, \bibinfo{person}{R. Saha~Roy},
  \bibinfo{person}{M. Yahya}, {and} \bibinfo{person}{G. Weikum}.}
  \bibinfo{year}{2018}\natexlab{}.
\newblock \showarticletitle{Never-ending learning for open-domain question
  answering over knowledge bases}. In \bibinfo{booktitle}{\emph{The WebConf}}.
  \bibinfo{pages}{1053--1062}.
\newblock


\bibitem[\protect\citeauthoryear{Battaglia, Hamrick, Bapst, Sanchez-Gonzalez,
  Zambaldi, Malinowski, et~al\mbox{.}}{Battaglia et~al\mbox{.}}{2018}]%
        {battaglia2018relational}
\bibfield{author}{\bibinfo{person}{P. Battaglia}, \bibinfo{person}{J.~B
  Hamrick}, \bibinfo{person}{V. Bapst}, \bibinfo{person}{A. Sanchez-Gonzalez},
  \bibinfo{person}{V. Zambaldi}, \bibinfo{person}{M. Malinowski},
  {et~al\mbox{.}}} \bibinfo{year}{2018}\natexlab{}.
\newblock \bibinfo{booktitle}{\emph{Relational inductive biases, deep learning,
  and graph networks}}.
\newblock \bibinfo{type}{{T}echnical {R}eport}.
  \bibinfo{institution}{arXiv:1806.01261}.
\newblock


\bibitem[\protect\citeauthoryear{Battista, Eades, Tamassia, and
  Tollis}{Battista et~al\mbox{.}}{1998}]%
        {battista1998graph}
\bibfield{author}{\bibinfo{person}{G.~D. Battista}, \bibinfo{person}{P. Eades},
  \bibinfo{person}{R. Tamassia}, {and} \bibinfo{person}{I.~G Tollis}.}
  \bibinfo{year}{1998}\natexlab{}.
\newblock \bibinfo{booktitle}{\emph{Graph drawing: algorithms for the
  visualization of graphs}}.
\newblock \bibinfo{publisher}{Prentice Hall PTR}.
\newblock


\bibitem[\protect\citeauthoryear{Berant and Liang}{Berant and Liang}{2014}]%
        {berant2014semantic}
\bibfield{author}{\bibinfo{person}{J. Berant} {and} \bibinfo{person}{P.
  Liang}.} \bibinfo{year}{2014}\natexlab{}.
\newblock \showarticletitle{Semantic parsing via paraphrasing}. In
  \bibinfo{booktitle}{\emph{ACL}}. \bibinfo{pages}{1415--1425}.
\newblock


\bibitem[\protect\citeauthoryear{Bordes, Usunier, Garcia-Duran, Weston, and
  Yakhnenko}{Bordes et~al\mbox{.}}{2013}]%
        {bordes2013translating}
\bibfield{author}{\bibinfo{person}{A. Bordes}, \bibinfo{person}{N. Usunier},
  \bibinfo{person}{A. Garcia-Duran}, \bibinfo{person}{J. Weston}, {and}
  \bibinfo{person}{O. Yakhnenko}.} \bibinfo{year}{2013}\natexlab{}.
\newblock \showarticletitle{Translating embeddings for modeling
  multi-relational data}. In \bibinfo{booktitle}{\emph{NeurIPS}}.
  \bibinfo{pages}{2787--2795}.
\newblock


\bibitem[\protect\citeauthoryear{Cao, Wang, He, Hu, and Chua}{Cao
  et~al\mbox{.}}{2019}]%
        {cao2019unifying}
\bibfield{author}{\bibinfo{person}{Y. Cao}, \bibinfo{person}{X. Wang},
  \bibinfo{person}{X. He}, \bibinfo{person}{Z. Hu}, {and} \bibinfo{person}{T.
  Chua}.} \bibinfo{year}{2019}\natexlab{}.
\newblock \showarticletitle{Unifying knowledge graph learning and
  recommendation: Towards a better understanding of user preferences}. In
  \bibinfo{booktitle}{\emph{The WebConf}}. \bibinfo{pages}{151--161}.
\newblock


\bibitem[\protect\citeauthoryear{Chen, Jia, and Xiang}{Chen
  et~al\mbox{.}}{2020}]%
        {chen2020review}
\bibfield{author}{\bibinfo{person}{X. Chen}, \bibinfo{person}{S. Jia}, {and}
  \bibinfo{person}{Y. Xiang}.} \bibinfo{year}{2020}\natexlab{}.
\newblock \showarticletitle{A review: Knowledge reasoning over knowledge
  graph}.
\newblock \bibinfo{journal}{\emph{Expert Systems with Applications}}
  \bibinfo{volume}{141} (\bibinfo{year}{2020}), \bibinfo{pages}{112948}.
\newblock


\bibitem[\protect\citeauthoryear{Cohen, Sun, Hofer, and Siegler}{Cohen
  et~al\mbox{.}}{2019}]%
        {cohen2019scalable}
\bibfield{author}{\bibinfo{person}{W.~W Cohen}, \bibinfo{person}{H. Sun},
  \bibinfo{person}{R~A. Hofer}, {and} \bibinfo{person}{M. Siegler}.}
  \bibinfo{year}{2019}\natexlab{}.
\newblock \showarticletitle{Scalable Neural Methods for Reasoning With a
  Symbolic Knowledge Base}. In \bibinfo{booktitle}{\emph{ICLR}}.
\newblock


\bibitem[\protect\citeauthoryear{Das, Dhuliawala, Zaheer, Vilnis, Durugkar,
  Krishnamurthy, Smola, and McCallum}{Das et~al\mbox{.}}{2017}]%
        {das2017go}
\bibfield{author}{\bibinfo{person}{R. Das}, \bibinfo{person}{S. Dhuliawala},
  \bibinfo{person}{M. Zaheer}, \bibinfo{person}{L. Vilnis}, \bibinfo{person}{I.
  Durugkar}, \bibinfo{person}{A. Krishnamurthy}, \bibinfo{person}{A. Smola},
  {and} \bibinfo{person}{A. McCallum}.} \bibinfo{year}{2017}\natexlab{}.
\newblock \showarticletitle{Go for a walk and arrive at the answer: Reasoning
  over paths in knowledge bases using reinforcement learning}. In
  \bibinfo{booktitle}{\emph{ICLR}}.
\newblock


\bibitem[\protect\citeauthoryear{Dettmers, Minervini, Stenetorp, and
  Riedel}{Dettmers et~al\mbox{.}}{2017}]%
        {dettmers2017convolutional}
\bibfield{author}{\bibinfo{person}{T. Dettmers}, \bibinfo{person}{P.
  Minervini}, \bibinfo{person}{P. Stenetorp}, {and} \bibinfo{person}{S.
  Riedel}.} \bibinfo{year}{2017}\natexlab{}.
\newblock \showarticletitle{Convolutional {2D} knowledge graph embeddings}. In
  \bibinfo{booktitle}{\emph{AAAI}}.
\newblock


\bibitem[\protect\citeauthoryear{Ding, Yao, Zhao, and Zhang}{Ding
  et~al\mbox{.}}{2021}]%
        {ding2021diffmg}
\bibfield{author}{\bibinfo{person}{Y. Ding}, \bibinfo{person}{Q. Yao},
  \bibinfo{person}{H. Zhao}, {and} \bibinfo{person}{T. Zhang}.}
  \bibinfo{year}{2021}\natexlab{}.
\newblock \showarticletitle{Diffmg: Differentiable meta graph search for
  heterogeneous graph neural networks}. In \bibinfo{booktitle}{\emph{SIGKDD}}.
  \bibinfo{pages}{279--288}.
\newblock


\bibitem[\protect\citeauthoryear{Gilmer, Schoenholz, Riley, Vinyals, and
  Dahl}{Gilmer et~al\mbox{.}}{2017}]%
        {gilmer2017neural}
\bibfield{author}{\bibinfo{person}{J. Gilmer}, \bibinfo{person}{S.~S
  Schoenholz}, \bibinfo{person}{P.~F Riley}, \bibinfo{person}{O. Vinyals},
  {and} \bibinfo{person}{G.~E Dahl}.} \bibinfo{year}{2017}\natexlab{}.
\newblock \showarticletitle{Neural Message Passing for Quantum Chemistry}. In
  \bibinfo{booktitle}{\emph{ICML}}. \bibinfo{pages}{1263--1272}.
\newblock


\bibitem[\protect\citeauthoryear{Grover and Leskovec}{Grover and
  Leskovec}{2016}]%
        {grover2016node2vec}
\bibfield{author}{\bibinfo{person}{A. Grover} {and} \bibinfo{person}{J.
  Leskovec}.} \bibinfo{year}{2016}\natexlab{}.
\newblock \showarticletitle{Node2vec: Scalable feature learning for networks}.
  In \bibinfo{booktitle}{\emph{SIGKDD}}. ACM, \bibinfo{pages}{855--864}.
\newblock


\bibitem[\protect\citeauthoryear{Hamilton, Ying, and Leskovec}{Hamilton
  et~al\mbox{.}}{2017}]%
        {hamilton2017inductive}
\bibfield{author}{\bibinfo{person}{W. Hamilton}, \bibinfo{person}{Z. Ying},
  {and} \bibinfo{person}{J. Leskovec}.} \bibinfo{year}{2017}\natexlab{}.
\newblock \showarticletitle{Inductive representation learning on large graphs}.
  In \bibinfo{booktitle}{\emph{NeurIPS}}. \bibinfo{pages}{1024--1034}.
\newblock


\bibitem[\protect\citeauthoryear{Harsha~V., Jia, and Kok}{Harsha~V.
  et~al\mbox{.}}{2020}]%
        {harsha2020probabilistic}
\bibfield{author}{\bibinfo{person}{L~V. Harsha~V.}, \bibinfo{person}{G. Jia},
  {and} \bibinfo{person}{S. Kok}.} \bibinfo{year}{2020}\natexlab{}.
\newblock \showarticletitle{Probabilistic logic graph attention networks for
  reasoning}. In \bibinfo{booktitle}{\emph{Companion of WebConf}}.
  \bibinfo{pages}{669--673}.
\newblock


\bibitem[\protect\citeauthoryear{Hogan, Blomqvist, Cochez, d'Amato, Melo,
  Gutierrez, Kirrane, Gayo, Navigli, Neumaier, et~al\mbox{.}}{Hogan
  et~al\mbox{.}}{2021}]%
        {hogan2021knowledge}
\bibfield{author}{\bibinfo{person}{A. Hogan}, \bibinfo{person}{E. Blomqvist},
  \bibinfo{person}{M. Cochez}, \bibinfo{person}{C. d'Amato},
  \bibinfo{person}{G.~D. Melo}, \bibinfo{person}{C. Gutierrez},
  \bibinfo{person}{S. Kirrane}, \bibinfo{person}{J.~E.~L. Gayo},
  \bibinfo{person}{R. Navigli}, \bibinfo{person}{S. Neumaier}, {et~al\mbox{.}}}
  \bibinfo{year}{2021}\natexlab{}.
\newblock \showarticletitle{Knowledge graphs}.
\newblock \bibinfo{journal}{\emph{CSUR}} \bibinfo{volume}{54},
  \bibinfo{number}{4} (\bibinfo{year}{2021}), \bibinfo{pages}{1--37}.
\newblock


\bibitem[\protect\citeauthoryear{Hornik}{Hornik}{1991}]%
        {hornik1991approximation}
\bibfield{author}{\bibinfo{person}{K. Hornik}.}
  \bibinfo{year}{1991}\natexlab{}.
\newblock \showarticletitle{Approximation capabilities of multilayer
  feedforward networks}.
\newblock \bibinfo{journal}{\emph{Neural networks}} \bibinfo{volume}{4},
  \bibinfo{number}{2} (\bibinfo{year}{1991}), \bibinfo{pages}{251--257}.
\newblock


\bibitem[\protect\citeauthoryear{Ji, Pan, Cambria, Marttinen, and Yu}{Ji
  et~al\mbox{.}}{2020}]%
        {ji2020survey}
\bibfield{author}{\bibinfo{person}{S. Ji}, \bibinfo{person}{S. Pan},
  \bibinfo{person}{E. Cambria}, \bibinfo{person}{P. Marttinen}, {and}
  \bibinfo{person}{P. Yu}.} \bibinfo{year}{2020}\natexlab{}.
\newblock \showarticletitle{A survey on knowledge graphs: Representation,
  acquisition and applications}.
\newblock \bibinfo{journal}{\emph{TKDE}} (\bibinfo{year}{2020}).
\newblock


\bibitem[\protect\citeauthoryear{Kingma and Ba}{Kingma and Ba}{2014}]%
        {kingma2014adam}
\bibfield{author}{\bibinfo{person}{D.~P Kingma} {and} \bibinfo{person}{J. Ba}.}
  \bibinfo{year}{2014}\natexlab{}.
\newblock \bibinfo{booktitle}{\emph{Adam: A method for stochastic
  optimization}}.
\newblock \bibinfo{type}{{T}echnical {R}eport}.
  \bibinfo{institution}{arXiv:1412.6980}.
\newblock


\bibitem[\protect\citeauthoryear{Kipf and Welling}{Kipf and Welling}{2016}]%
        {kipf2016semi}
\bibfield{author}{\bibinfo{person}{T. Kipf} {and} \bibinfo{person}{M.
  Welling}.} \bibinfo{year}{2016}\natexlab{}.
\newblock \showarticletitle{Semi-supervised classification with graph
  convolutional networks}. In \bibinfo{booktitle}{\emph{ICLR}}.
\newblock


\bibitem[\protect\citeauthoryear{Kok and Domingos}{Kok and Domingos}{2007}]%
        {kok2007statistical}
\bibfield{author}{\bibinfo{person}{S. Kok} {and} \bibinfo{person}{P.
  Domingos}.} \bibinfo{year}{2007}\natexlab{}.
\newblock \showarticletitle{Statistical predicate invention}. In
  \bibinfo{booktitle}{\emph{ICML}}. \bibinfo{pages}{433--440}.
\newblock


\bibitem[\protect\citeauthoryear{Lacroix, Usunier, and Obozinski}{Lacroix
  et~al\mbox{.}}{2018}]%
        {lacroix2018canonical}
\bibfield{author}{\bibinfo{person}{T. Lacroix}, \bibinfo{person}{N. Usunier},
  {and} \bibinfo{person}{G. Obozinski}.} \bibinfo{year}{2018}\natexlab{}.
\newblock \showarticletitle{Canonical Tensor Decomposition for Knowledge Base
  Completion}. In \bibinfo{booktitle}{\emph{ICML}}.
  \bibinfo{pages}{2863--2872}.
\newblock


\bibitem[\protect\citeauthoryear{Lao and Cohen}{Lao and Cohen}{2010}]%
        {lao2010relational}
\bibfield{author}{\bibinfo{person}{N. Lao} {and} \bibinfo{person}{W.~W Cohen}.}
  \bibinfo{year}{2010}\natexlab{}.
\newblock \showarticletitle{Relational retrieval using a combination of
  path-constrained random walks}.
\newblock \bibinfo{journal}{\emph{Machine learning}} \bibinfo{volume}{81},
  \bibinfo{number}{1} (\bibinfo{year}{2010}), \bibinfo{pages}{53--67}.
\newblock


\bibitem[\protect\citeauthoryear{Lao, Mitchell, and Cohen}{Lao
  et~al\mbox{.}}{2011}]%
        {lao2011random}
\bibfield{author}{\bibinfo{person}{N. Lao}, \bibinfo{person}{T. Mitchell},
  {and} \bibinfo{person}{W. Cohen}.} \bibinfo{year}{2011}\natexlab{}.
\newblock \showarticletitle{Random walk inference and learning in a large scale
  knowledge base}. In \bibinfo{booktitle}{\emph{EMNLP}}.
  \bibinfo{pages}{529--539}.
\newblock


\bibitem[\protect\citeauthoryear{Meilicke, Fink, Wang, Ruffinelli, Gemulla, and
  Stuckenschmidt}{Meilicke et~al\mbox{.}}{2018}]%
        {meilicke2018fine}
\bibfield{author}{\bibinfo{person}{C. Meilicke}, \bibinfo{person}{M. Fink},
  \bibinfo{person}{Y. Wang}, \bibinfo{person}{D. Ruffinelli},
  \bibinfo{person}{R. Gemulla}, {and} \bibinfo{person}{H. Stuckenschmidt}.}
  \bibinfo{year}{2018}\natexlab{}.
\newblock \showarticletitle{Fine-grained evaluation of rule-and embedding-based
  systems for knowledge graph completion}. In \bibinfo{booktitle}{\emph{ISWC}}.
  Springer, \bibinfo{pages}{3--20}.
\newblock


\bibitem[\protect\citeauthoryear{Niepert}{Niepert}{2016}]%
        {niepert2016discriminative}
\bibfield{author}{\bibinfo{person}{M. Niepert}.}
  \bibinfo{year}{2016}\natexlab{}.
\newblock \showarticletitle{Discriminative gaifman models}. In
  \bibinfo{booktitle}{\emph{NIPS}}, Vol.~\bibinfo{volume}{29}.
  \bibinfo{pages}{3405--3413}.
\newblock


\bibitem[\protect\citeauthoryear{Paszke, Gross, Chintala, Chanan, Yang, DeVito,
  Lin, Desmaison, Antiga, and Lerer}{Paszke et~al\mbox{.}}{2017}]%
        {paszke2017automatic}
\bibfield{author}{\bibinfo{person}{A. Paszke}, \bibinfo{person}{S. Gross},
  \bibinfo{person}{S. Chintala}, \bibinfo{person}{G. Chanan},
  \bibinfo{person}{E. Yang}, \bibinfo{person}{Z. DeVito}, \bibinfo{person}{Z.
  Lin}, \bibinfo{person}{A. Desmaison}, \bibinfo{person}{L. Antiga}, {and}
  \bibinfo{person}{A. Lerer}.} \bibinfo{year}{2017}\natexlab{}.
\newblock \showarticletitle{Automatic differentiation in {PyTorch}}. In
  \bibinfo{booktitle}{\emph{ICLR}}.
\newblock


\bibitem[\protect\citeauthoryear{Perozzi, Al-Rfou, and Skiena}{Perozzi
  et~al\mbox{.}}{2014}]%
        {perozzi2014deepwalk}
\bibfield{author}{\bibinfo{person}{B. Perozzi}, \bibinfo{person}{R. Al-Rfou},
  {and} \bibinfo{person}{S. Skiena}.} \bibinfo{year}{2014}\natexlab{}.
\newblock \showarticletitle{Deepwalk: Online learning of social
  representations}. In \bibinfo{booktitle}{\emph{SIGKDD}}. ACM,
  \bibinfo{pages}{701--710}.
\newblock


\bibitem[\protect\citeauthoryear{Qu, Chen, Xhonneux, Bengio, and Tang}{Qu
  et~al\mbox{.}}{2021}]%
        {qu2021rnnlogic}
\bibfield{author}{\bibinfo{person}{M. Qu}, \bibinfo{person}{J. Chen},
  \bibinfo{person}{L. Xhonneux}, \bibinfo{person}{Y. Bengio}, {and}
  \bibinfo{person}{J. Tang}.} \bibinfo{year}{2021}\natexlab{}.
\newblock \showarticletitle{RNNLogic: Learning Logic Rules for Reasoning on
  Knowledge Graphs}. In \bibinfo{booktitle}{\emph{ICLR}}.
\newblock


\bibitem[\protect\citeauthoryear{Qu and Tang}{Qu and Tang}{2019}]%
        {qu2019probabilistic}
\bibfield{author}{\bibinfo{person}{M. Qu} {and} \bibinfo{person}{J. Tang}.}
  \bibinfo{year}{2019}\natexlab{}.
\newblock \showarticletitle{Probabilistic Logic Neural Networks for Reasoning}.
\newblock \bibinfo{journal}{\emph{NeurIPS}}  \bibinfo{volume}{32},
  \bibinfo{pages}{7712--7722}.
\newblock


\bibitem[\protect\citeauthoryear{Rossi, Firmani, Matinata, Merialdo, and
  Barbosa}{Rossi et~al\mbox{.}}{2021}]%
        {rossi2020knowledge}
\bibfield{author}{\bibinfo{person}{A. Rossi}, \bibinfo{person}{D. Firmani},
  \bibinfo{person}{A. Matinata}, \bibinfo{person}{P. Merialdo}, {and}
  \bibinfo{person}{D. Barbosa}.} \bibinfo{year}{2021}\natexlab{}.
\newblock \showarticletitle{Knowledge Graph Embedding for Link Prediction: A
  Comparative Analysis}.
\newblock \bibinfo{journal}{\emph{TKDD}} (\bibinfo{year}{2021}).
\newblock


\bibitem[\protect\citeauthoryear{Ruffinelli, Broscheit, and Gemulla}{Ruffinelli
  et~al\mbox{.}}{2020}]%
        {ruffinelli2020you}
\bibfield{author}{\bibinfo{person}{D. Ruffinelli}, \bibinfo{person}{S.
  Broscheit}, {and} \bibinfo{person}{R. Gemulla}.}
  \bibinfo{year}{2020}\natexlab{}.
\newblock \showarticletitle{You can teach an old dog new tricks! on training
  knowledge graph embeddings}. In \bibinfo{booktitle}{\emph{ICLR}}.
\newblock


\bibitem[\protect\citeauthoryear{Sadeghian, Armandpour, Ding, and
  Wang}{Sadeghian et~al\mbox{.}}{2019}]%
        {sadeghian2019drum}
\bibfield{author}{\bibinfo{person}{A. Sadeghian}, \bibinfo{person}{M.
  Armandpour}, \bibinfo{person}{P. Ding}, {and} \bibinfo{person}{D Wang}.}
  \bibinfo{year}{2019}\natexlab{}.
\newblock \showarticletitle{DRUM: End-To-End Differentiable Rule Mining On
  Knowledge Graphs}. In \bibinfo{booktitle}{\emph{NeurIPS}}.
  \bibinfo{pages}{15347--15357}.
\newblock


\bibitem[\protect\citeauthoryear{Schlichtkrull, Kipf, Bloem, Van~D., Titov, and
  Welling}{Schlichtkrull et~al\mbox{.}}{2018}]%
        {schlichtkrull2018modeling}
\bibfield{author}{\bibinfo{person}{M. Schlichtkrull}, \bibinfo{person}{T.~N
  Kipf}, \bibinfo{person}{P. Bloem}, \bibinfo{person}{Rianne Van~D.},
  \bibinfo{person}{I. Titov}, {and} \bibinfo{person}{M. Welling}.}
  \bibinfo{year}{2018}\natexlab{}.
\newblock \showarticletitle{Modeling relational data with graph convolutional
  networks}. In \bibinfo{booktitle}{\emph{ESWC}}. Springer,
  \bibinfo{pages}{593--607}.
\newblock


\bibitem[\protect\citeauthoryear{Shen, Chen, Huang, and Gao}{Shen
  et~al\mbox{.}}{2018}]%
        {shen2018m}
\bibfield{author}{\bibinfo{person}{Y. Shen}, \bibinfo{person}{J. Chen},
  \bibinfo{person}{Y. Huang, P.and~Guo}, {and} \bibinfo{person}{J. Gao}.}
  \bibinfo{year}{2018}\natexlab{}.
\newblock \showarticletitle{M-walk: Learning to walk over graphs using monte
  carlo tree search}. In \bibinfo{booktitle}{\emph{NeurIPS}}.
\newblock


\bibitem[\protect\citeauthoryear{Sun, Deng, Nie, and Tang}{Sun
  et~al\mbox{.}}{2019}]%
        {sun2019rotate}
\bibfield{author}{\bibinfo{person}{Z. Sun}, \bibinfo{person}{Z. Deng},
  \bibinfo{person}{J. Nie}, {and} \bibinfo{person}{J. Tang}.}
  \bibinfo{year}{2019}\natexlab{}.
\newblock \showarticletitle{{RotatE}: Knowledge graph embedding by relational
  rotation in complex space}. In \bibinfo{booktitle}{\emph{ICLR}}.
\newblock


\bibitem[\protect\citeauthoryear{Sun, Vashishth, Sanyal, Talukdar, and
  Yang}{Sun et~al\mbox{.}}{2020}]%
        {sun2020re}
\bibfield{author}{\bibinfo{person}{Z. Sun}, \bibinfo{person}{S. Vashishth},
  \bibinfo{person}{S. Sanyal}, \bibinfo{person}{P. Talukdar}, {and}
  \bibinfo{person}{Y. Yang}.} \bibinfo{year}{2020}\natexlab{}.
\newblock \showarticletitle{A Re-evaluation of Knowledge Graph Completion
  Methods}. In \bibinfo{booktitle}{\emph{ACL}}. \bibinfo{pages}{5516--5522}.
\newblock


\bibitem[\protect\citeauthoryear{Teru, Denis, and Hamilton}{Teru
  et~al\mbox{.}}{2020}]%
        {teru2019inductive}
\bibfield{author}{\bibinfo{person}{K.~K Teru}, \bibinfo{person}{E. Denis},
  {and} \bibinfo{person}{W. Hamilton}.} \bibinfo{year}{2020}\natexlab{}.
\newblock \showarticletitle{Inductive Relation Prediction by Subgraph
  Reasoning}. In \bibinfo{booktitle}{\emph{ICML}}.
\newblock


\bibitem[\protect\citeauthoryear{Toutanova and Chen}{Toutanova and
  Chen}{2015}]%
        {toutanova2015observed}
\bibfield{author}{\bibinfo{person}{K. Toutanova} {and} \bibinfo{person}{D.
  Chen}.} \bibinfo{year}{2015}\natexlab{}.
\newblock \showarticletitle{Observed versus latent features for knowledge base
  and text inference}. In \bibinfo{booktitle}{\emph{PWCVSMC}}.
  \bibinfo{pages}{57--66}.
\newblock


\bibitem[\protect\citeauthoryear{Trouillon, Dance, Gaussier, Welbl, Riedel, and
  Bouchard}{Trouillon et~al\mbox{.}}{2017}]%
        {trouillon2017knowledge}
\bibfield{author}{\bibinfo{person}{T. Trouillon}, \bibinfo{person}{C.~R Dance},
  \bibinfo{person}{{\'E}. Gaussier}, \bibinfo{person}{J. Welbl},
  \bibinfo{person}{S. Riedel}, {and} \bibinfo{person}{G. Bouchard}.}
  \bibinfo{year}{2017}\natexlab{}.
\newblock \showarticletitle{Knowledge graph completion via complex tensor
  factorization}.
\newblock \bibinfo{journal}{\emph{JMLR}} \bibinfo{volume}{18},
  \bibinfo{number}{1} (\bibinfo{year}{2017}), \bibinfo{pages}{4735--4772}.
\newblock


\bibitem[\protect\citeauthoryear{Vashishth, Sanyal, Nitin, and
  Talukdar}{Vashishth et~al\mbox{.}}{2019}]%
        {vashishth2019composition}
\bibfield{author}{\bibinfo{person}{S. Vashishth}, \bibinfo{person}{S. Sanyal},
  \bibinfo{person}{V. Nitin}, {and} \bibinfo{person}{P. Talukdar}.}
  \bibinfo{year}{2019}\natexlab{}.
\newblock \showarticletitle{Composition-based multi-relational graph
  convolutional networks}. In \bibinfo{booktitle}{\emph{ICLR}}.
\newblock


\bibitem[\protect\citeauthoryear{Veli{\v{c}}kovi{\'c}, Cucurull, Casanova,
  Romero, Lio, and Bengio}{Veli{\v{c}}kovi{\'c} et~al\mbox{.}}{2017}]%
        {velivckovic2017graph}
\bibfield{author}{\bibinfo{person}{P. Veli{\v{c}}kovi{\'c}},
  \bibinfo{person}{G. Cucurull}, \bibinfo{person}{A. Casanova},
  \bibinfo{person}{A. Romero}, \bibinfo{person}{P. Lio}, {and}
  \bibinfo{person}{Y. Bengio}.} \bibinfo{year}{2017}\natexlab{}.
\newblock \showarticletitle{Graph attention networks}. In
  \bibinfo{booktitle}{\emph{ICLR}}.
\newblock


\bibitem[\protect\citeauthoryear{Wang, Ren, and Leskovec}{Wang
  et~al\mbox{.}}{2021}]%
        {wang2021relational}
\bibfield{author}{\bibinfo{person}{H. Wang}, \bibinfo{person}{H. Ren}, {and}
  \bibinfo{person}{J. Leskovec}.} \bibinfo{year}{2021}\natexlab{}.
\newblock \showarticletitle{Relational Message Passing for Knowledge Graph
  Completion}. In \bibinfo{booktitle}{\emph{SIGKDD}}.
  \bibinfo{pages}{1697--1707}.
\newblock


\bibitem[\protect\citeauthoryear{Wang, Mao, Wang, and Guo}{Wang
  et~al\mbox{.}}{2017}]%
        {wang2017knowledge}
\bibfield{author}{\bibinfo{person}{Q. Wang}, \bibinfo{person}{Z. Mao},
  \bibinfo{person}{B. Wang}, {and} \bibinfo{person}{L. Guo}.}
  \bibinfo{year}{2017}\natexlab{}.
\newblock \showarticletitle{Knowledge graph embedding: A survey of approaches
  and applications}.
\newblock \bibinfo{journal}{\emph{TKDE}} \bibinfo{volume}{29},
  \bibinfo{number}{12} (\bibinfo{year}{2017}), \bibinfo{pages}{2724--2743}.
\newblock


\bibitem[\protect\citeauthoryear{Xiao, Zhao, Zheng, and Song}{Xiao
  et~al\mbox{.}}{2021}]%
        {xiao2021neural}
\bibfield{author}{\bibinfo{person}{W. Xiao}, \bibinfo{person}{H. Zhao},
  \bibinfo{person}{V.~W Zheng}, {and} \bibinfo{person}{Y. Song}.}
  \bibinfo{year}{2021}\natexlab{}.
\newblock \showarticletitle{Neural PathSim for Inductive Similarity Search in
  Heterogeneous Information Networks}. In \bibinfo{booktitle}{\emph{CIKM}}.
  \bibinfo{pages}{2201--2210}.
\newblock


\bibitem[\protect\citeauthoryear{Xiong, Hoang, and Wang}{Xiong
  et~al\mbox{.}}{2017}]%
        {xiong2017deeppath}
\bibfield{author}{\bibinfo{person}{W. Xiong}, \bibinfo{person}{T. Hoang}, {and}
  \bibinfo{person}{W. Wang}.} \bibinfo{year}{2017}\natexlab{}.
\newblock \showarticletitle{DeepPath: A Reinforcement Learning Method for
  Knowledge Graph Reasoning}. In \bibinfo{booktitle}{\emph{EMNLP}}.
  \bibinfo{pages}{564--573}.
\newblock


\bibitem[\protect\citeauthoryear{Xu, Feng, Jiang, Xie, Sun, and Deng}{Xu
  et~al\mbox{.}}{2019}]%
        {xu2019dynamically}
\bibfield{author}{\bibinfo{person}{X. Xu}, \bibinfo{person}{W. Feng},
  \bibinfo{person}{Y. Jiang}, \bibinfo{person}{X. Xie}, \bibinfo{person}{Z.
  Sun}, {and} \bibinfo{person}{Z. Deng}.} \bibinfo{year}{2019}\natexlab{}.
\newblock \showarticletitle{Dynamically Pruned Message Passing Networks for
  Large-Scale Knowledge Graph Reasoning}. In \bibinfo{booktitle}{\emph{ICLR}}.
\newblock


\bibitem[\protect\citeauthoryear{Yang, Yang, and Cohen}{Yang
  et~al\mbox{.}}{2017}]%
        {yang2017differentiable}
\bibfield{author}{\bibinfo{person}{F. Yang}, \bibinfo{person}{Z. Yang}, {and}
  \bibinfo{person}{W. Cohen}.} \bibinfo{year}{2017}\natexlab{}.
\newblock \showarticletitle{Differentiable learning of logical rules for
  knowledge base reasoning}. In \bibinfo{booktitle}{\emph{NeurIPS}}.
  \bibinfo{pages}{2319--2328}.
\newblock


\bibitem[\protect\citeauthoryear{Yu, Yang, Zhang, and Wu}{Yu
  et~al\mbox{.}}{2021}]%
        {yu2021knowledge}
\bibfield{author}{\bibinfo{person}{D. Yu}, \bibinfo{person}{Y. Yang},
  \bibinfo{person}{R. Zhang}, {and} \bibinfo{person}{Y. Wu}.}
  \bibinfo{year}{2021}\natexlab{}.
\newblock \showarticletitle{Knowledge Embedding Based Graph Convolutional
  Network}. In \bibinfo{booktitle}{\emph{The WebConf}}.
  \bibinfo{pages}{1619--1628}.
\newblock


\bibitem[\protect\citeauthoryear{Zhang and Chen}{Zhang and Chen}{2019}]%
        {zhang2019inductive}
\bibfield{author}{\bibinfo{person}{M. Zhang} {and} \bibinfo{person}{Y. Chen}.}
  \bibinfo{year}{2019}\natexlab{}.
\newblock \showarticletitle{Inductive Matrix Completion Based on Graph Neural
  Networks}. In \bibinfo{booktitle}{\emph{ICLR}}.
\newblock


\bibitem[\protect\citeauthoryear{Zhang, Tay, Yao, and Liu}{Zhang
  et~al\mbox{.}}{2019a}]%
        {zhang2019quaternion}
\bibfield{author}{\bibinfo{person}{S. Zhang}, \bibinfo{person}{Y. Tay},
  \bibinfo{person}{L. Yao}, {and} \bibinfo{person}{Q. Liu}.}
  \bibinfo{year}{2019}\natexlab{a}.
\newblock \showarticletitle{Quaternion knowledge graph embeddings}. In
  \bibinfo{booktitle}{\emph{NeurIPS}}.
\newblock


\bibitem[\protect\citeauthoryear{Zhang, Dai, Kozareva, Smola, and Song}{Zhang
  et~al\mbox{.}}{2018}]%
        {zhang2018variational}
\bibfield{author}{\bibinfo{person}{Y. Zhang}, \bibinfo{person}{H. Dai},
  \bibinfo{person}{Z. Kozareva}, \bibinfo{person}{A.~J Smola}, {and}
  \bibinfo{person}{L. Song}.} \bibinfo{year}{2018}\natexlab{}.
\newblock \showarticletitle{Variational reasoning for question answering with
  knowledge graph}. In \bibinfo{booktitle}{\emph{AAAI}}.
\newblock


\bibitem[\protect\citeauthoryear{Zhang, Yao, and Chen}{Zhang
  et~al\mbox{.}}{2020a}]%
        {zhang2020interstellar}
\bibfield{author}{\bibinfo{person}{Y. Zhang}, \bibinfo{person}{Q. Yao}, {and}
  \bibinfo{person}{L. Chen}.} \bibinfo{year}{2020}\natexlab{a}.
\newblock \showarticletitle{Interstellar: Searching Recurrent Architecture for
  Knowledge Graph Embedding}. In \bibinfo{booktitle}{\emph{NeurIPS}},
  Vol.~\bibinfo{volume}{33}.
\newblock


\bibitem[\protect\citeauthoryear{Zhang, Yao, Dai, and Chen}{Zhang
  et~al\mbox{.}}{2020b}]%
        {zhang2020autosf}
\bibfield{author}{\bibinfo{person}{Y. Zhang}, \bibinfo{person}{Q. Yao},
  \bibinfo{person}{W. Dai}, {and} \bibinfo{person}{L. Chen}.}
  \bibinfo{year}{2020}\natexlab{b}.
\newblock \showarticletitle{AutoSF: Searching scoring functions for knowledge
  graph embedding}. In \bibinfo{booktitle}{\emph{ICDE}}. IEEE,
  \bibinfo{pages}{433--444}.
\newblock


\bibitem[\protect\citeauthoryear{Zhang, Yao, Shao, and Chen}{Zhang
  et~al\mbox{.}}{2019b}]%
        {zhang2019nscaching}
\bibfield{author}{\bibinfo{person}{Y. Zhang}, \bibinfo{person}{Q. Yao},
  \bibinfo{person}{Y. Shao}, {and} \bibinfo{person}{L. Chen}.}
  \bibinfo{year}{2019}\natexlab{b}.
\newblock \showarticletitle{{NSCaching}: simple and efficient negative sampling
  for knowledge graph embedding}. In \bibinfo{booktitle}{\emph{ICDE}}. IEEE,
  \bibinfo{pages}{614--625}.
\newblock


\bibitem[\protect\citeauthoryear{Zhao, Yao, Li, Song, and Lee}{Zhao
  et~al\mbox{.}}{2017}]%
        {zhao2017meta}
\bibfield{author}{\bibinfo{person}{H. Zhao}, \bibinfo{person}{Q. Yao},
  \bibinfo{person}{J. Li}, \bibinfo{person}{Y. Song}, {and}
  \bibinfo{person}{K.~L. Lee}.} \bibinfo{year}{2017}\natexlab{}.
\newblock \showarticletitle{Meta-graph based recommendation fusion over
  heterogeneous information networks}. In \bibinfo{booktitle}{\emph{SIGKDD}}.
  \bibinfo{pages}{635--644}.
\newblock


\bibitem[\protect\citeauthoryear{Zhu, Zhang, Xhonneux, and Tang}{Zhu
  et~al\mbox{.}}{2021}]%
        {zhu2021neural}
\bibfield{author}{\bibinfo{person}{Z. Zhu}, \bibinfo{person}{Z. Zhang},
  \bibinfo{person}{L. Xhonneux}, {and} \bibinfo{person}{J. Tang}.}
  \bibinfo{year}{2021}\natexlab{}.
\newblock \showarticletitle{Neural {Bellman-Ford} Networks: A General Graph
  Neural Network Framework for Link Prediction}, In
  \bibinfo{booktitle}{NeurIPS}.
\newblock \bibinfo{journal}{\emph{arXiv e-prints}},
  \bibinfo{pages}{arXiv--2106}.
\newblock


\end{thebibliography}

\clearpage
\appendix

\section{Visualization}
\label{app:visual}

We provide the visualization of the r-digraph learned between entities in Algorithm~\ref{alg:vis}.
The key point is
to backtrack the neighborhood edges of $e_a$
that has attention weight larger than a pre-defined threshold $\theta>0$,
i.e., steps~3-6.
The remaining structures 
$\mathcal G_{e_q,e_a|L}(\theta)$
are returned as the structure selected by the attention weights.

\begin{algorithm}[ht]
	\caption{Visualization.}
	\small
	\begin{algorithmic}[1]
		\REQUIRE entities $\mathcal V$, triples $\mathcal F$, query triple $(e_q, r_q, e_a)$, depth $L$, 
		parameters $\bm \Theta$, threshold $\theta$.
		\STATE Run Algorithm~\ref{alg:gnn} to obtain the attention weights $\alpha_{e_s,\!r,\!e_o\!|r_q}^{\ell}, \ell=1\dots L$ 
		on the edges in each layer.
		\STATE Initialize ${\mathcal  V}_{e_q, e_a|L}^L(\theta) = \{e_a\}$.
		\FOR{$\ell=L, L-1 \dots 1$}
		\STATE collect the edges ${\mathcal E}_{e_q, e_a|L}^{\ell}(\theta) \!=\! \{(e_s, r, e_o)|e_o\!\in\!{\mathcal  V}_{e_q, e_a|L}^L , \alpha_{e_s,\!r,\!e_o\!|r_q}^{\ell}\!\geq\!\theta\}$.
		\STATE collect the entities ${\mathcal V}_{e_q, e_a|L}^{\ell-1}(\theta) =\{e_s| (e_s, r, e_o)\in {\mathcal E}_{e_q, e_a|L}^{\ell}(\theta)  \}$.
		\ENDFOR
		
		\RETURN $\mathcal G_{e_q,e_a|L}(\theta) = {\mathcal E}_{e_q, e_a|L}^{1}(\theta)\otimes{\mathcal E}_{e_q, e_a|L}^{2}(\theta)\cdots \otimes {\mathcal E}_{e_q, e_a|L}^{L}(\theta).$
	\end{algorithmic}
	\label{alg:vis}
\end{algorithm}

\begin{figure}[ht]
	\centering
	{\includegraphics[width=0.9\columnwidth]{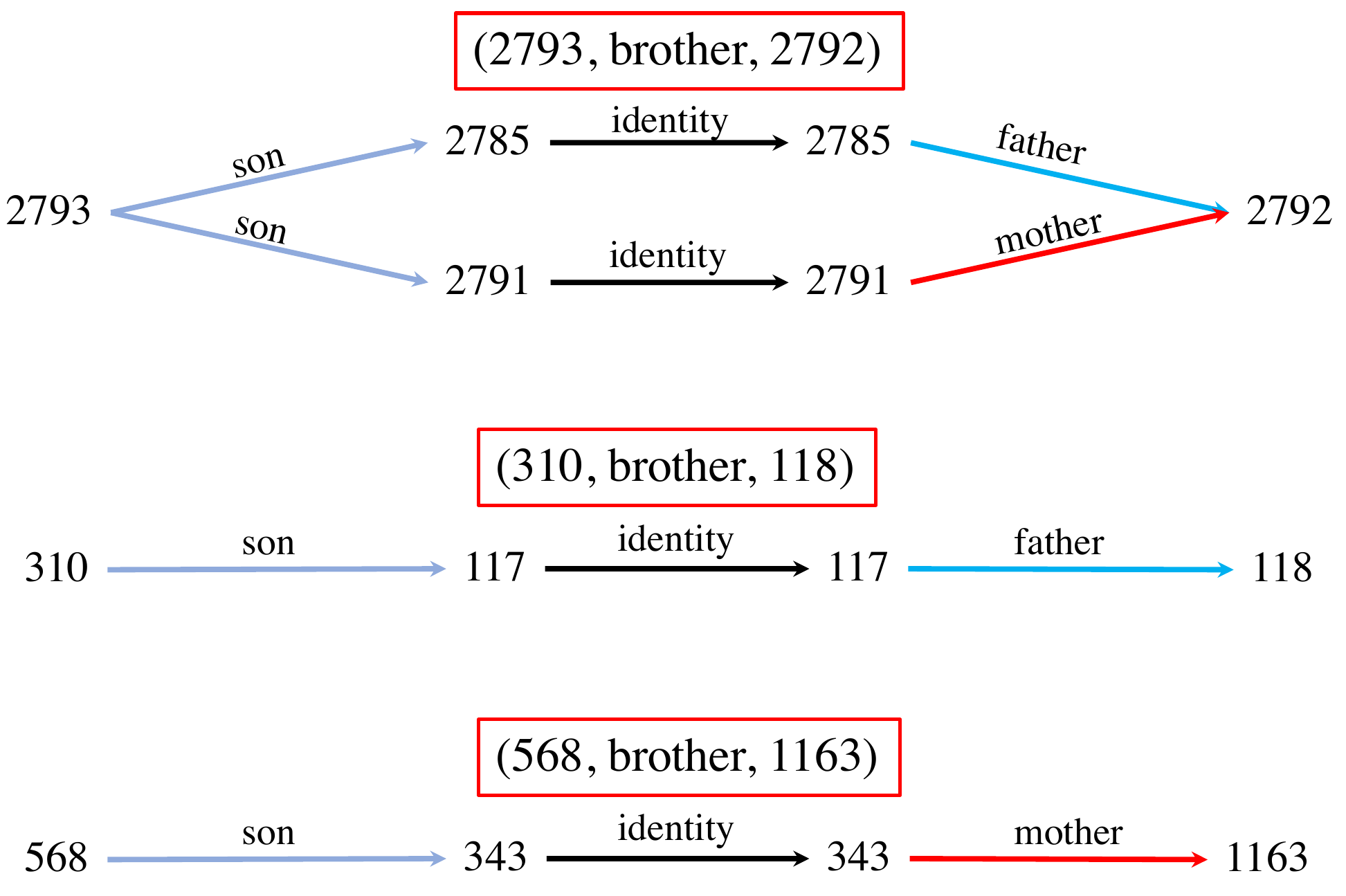}}
	\vspace{-8px}
	\caption{Visualization of the learned structure on Family.}
	\label{fig:app:family}
\end{figure}

\begin{figure}[ht]
	\centering
	{\includegraphics[width=0.8\columnwidth]{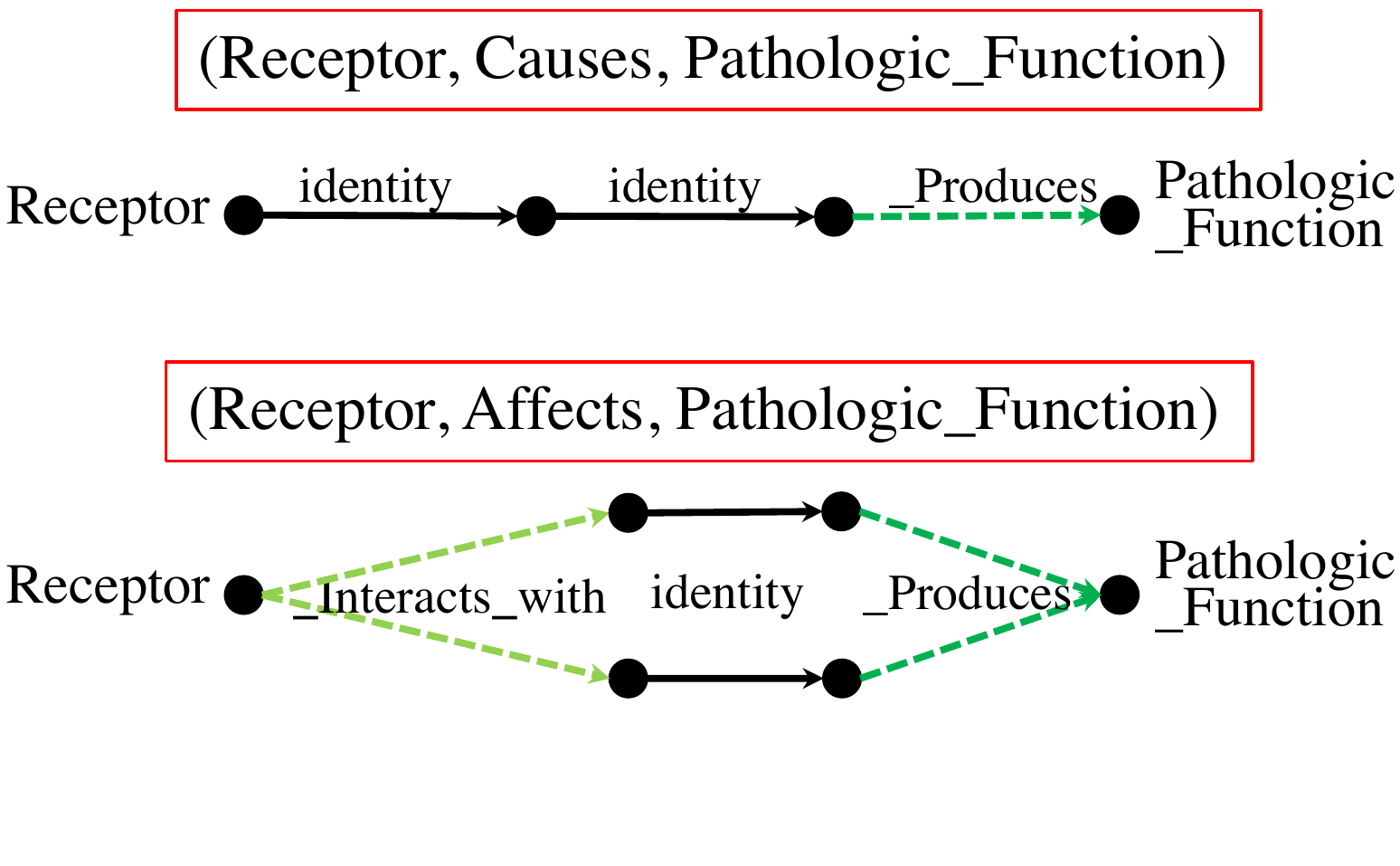}\vspace{-15px}}
	{\includegraphics[width=1.0\columnwidth]{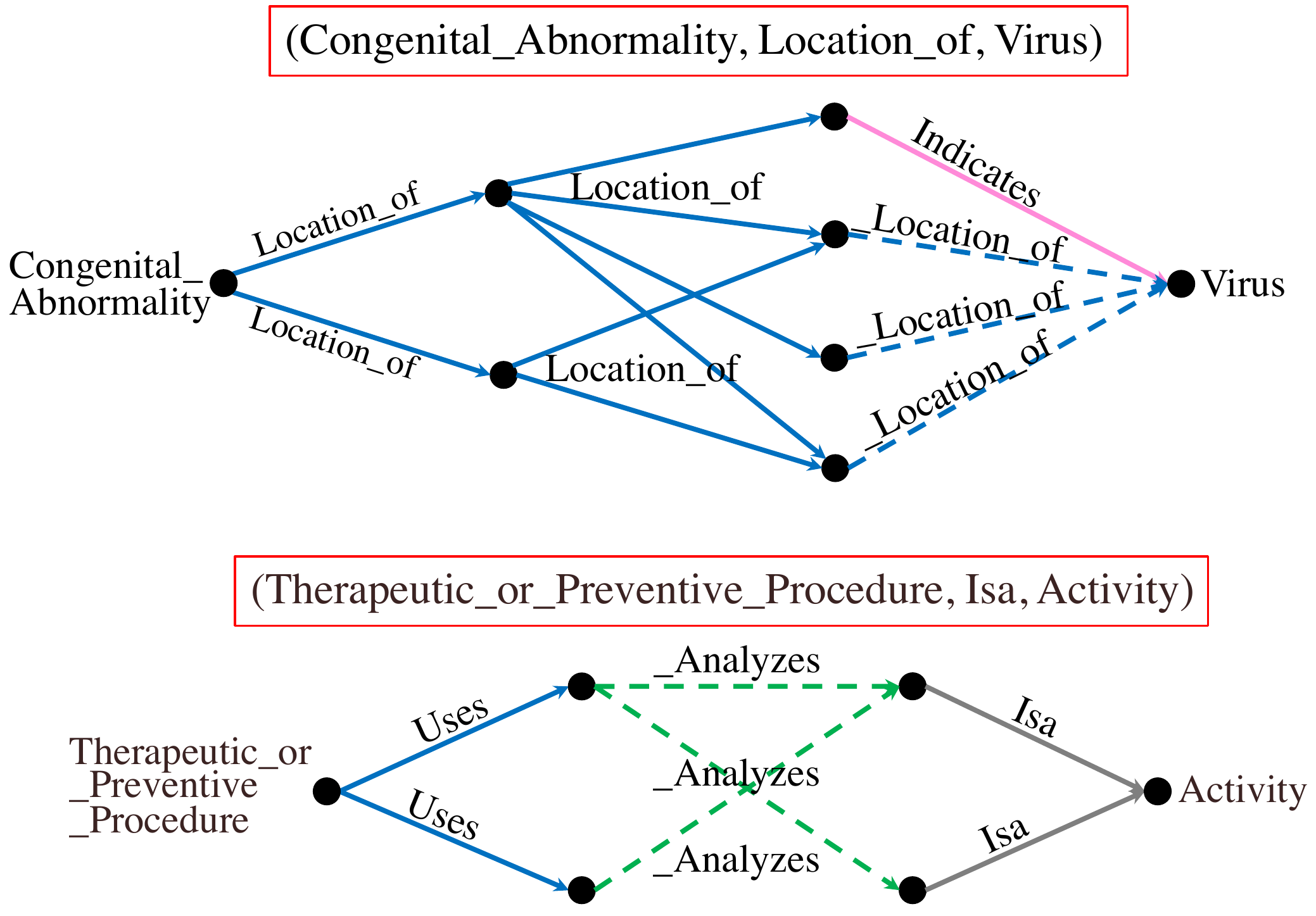}}
	\vspace{-8px}
	\caption{Visualization of the learned structure on UMLS. }
	\label{fig:app:umls}
\end{figure}

We provide some additional visualized results on Family and UMLS datasets
in Figure~~\ref{fig:app:family} and Figure~~\ref{fig:app:umls},.
There are consistent and semantically friendly patterns across different query triples.

\section{Problem analysis on GraIL}
\label{app:grail}

As mentioned in the main text,
efficiency is the preliminary issue of GraIL~\cite{teru2019inductive}.
In this part,
we mainly analyze it in terms of effectiveness.


\noindent
\textbf{Enclosing subgraph in \cite{teru2019inductive} v.s. r-digraph.}
When constructing the enclosing subgraph,
the bidirectional information between $e_q$ and $e_a$ is preserved.
It contains all the relational paths from $e_q$ to $e_a$
and all the paths from $e_a$ to $e_q$.
This is the key difference between enclosing subgraph and r-digraph.
In this view,
the limitations of the enclosing subgraph can be summarized as follows.
\begin{itemize}[leftmargin=*]
	\item 
	The entities in the enclosing subgraph need distance as labels.
	But in r-digraph, the distance is implicitly encoded in layers.
	
	\item The order of relations, inside the enclosing subgraph is mixed,
	making it hard to learn the order of relations,
	e.g. the difference between 
	\textit{brother}$\wedge$\textit{mother}$\rightarrow$\textit{uncle} 
	and \textit{mother}$\wedge$\textit{brother}$\rightarrow$\textit{aunt}.
\end{itemize}

The enclosing subgraph, even with more edges than the r-digraph,
does not show valuable inductive bias for KG reasoning.

\begin{figure}[H]
	\centering
	\includegraphics[width=0.9\columnwidth]{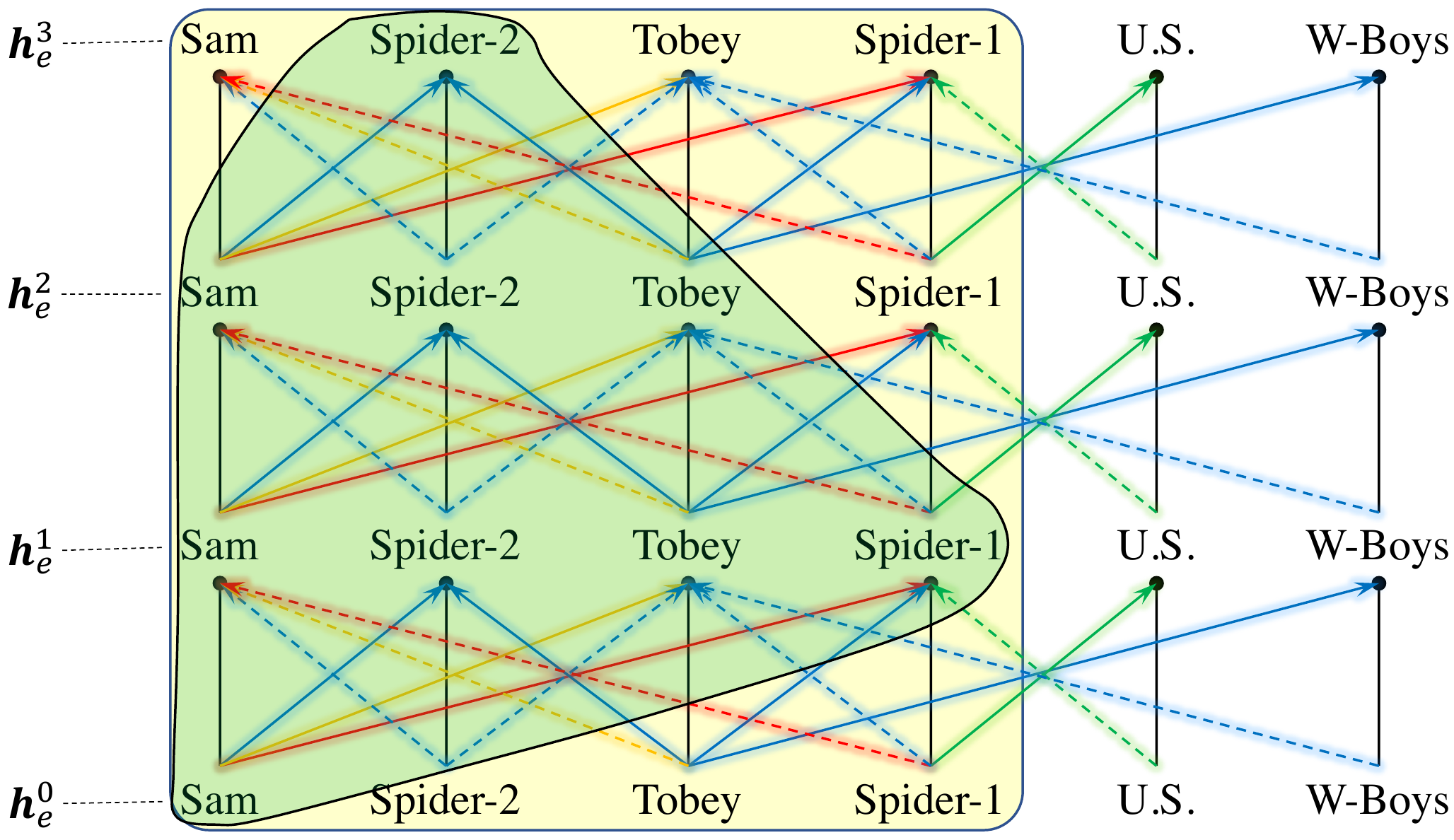}
	\vspace{-10px}
	\caption{The computation graph of GraIL (yellow space) and RED-GNN (green space).}
	\label{fig:grail}
\end{figure}

\noindent
\textbf{Non-interpretable.}
We show the computation graphs of GraIL and RED-GNN
in the yellow space and green space respectively
in Figure~\ref{fig:grail}.
Even though attention is applied in the GNN framework,
how the weights cooperate in different layers is unclear.
Thus they did not provide interpretation
and nor can we come up a way to interpret 
the reasoning results in GraIL.

\noindent
\textbf{Empirical evidence.}
To show that GraIL fails to learn the relational structures,
we conduct a tiny ablation study here.
Specifically,
we change the message function in GraIL
from
\[\bm a_t^k = \sum\nolimits_{r=1}^R \sum\nolimits_{s\in\mathcal N_r(t)} \alpha_{r,r_t,s,t}^k\bm W_r^k\bm h_s^{k-1},\]
\[
\text{to } ~\bm a_t^k = \sum\nolimits_{r=1}^R \sum\nolimits_{s\in\mathcal N_r(t)} \alpha_{r_t,s,t}^k\bm W^k\bm h_s^{k-1}, \]
where $\alpha_{r_t,s,t}^k = \text{MLP}(\bm h_s^{k-1}, \bm h_t^{k-1}, \bm e_{r_t}^a)$.
This removes the relation when aggregating edges to only focus on graph structures.

We name this as GraIL no rel.
The results are shown in Table~\ref{tab:homo}.
We observe that the performance decay is really marginal
if relation is removed in the edges.
This means that GraIL mainly learns on
 the graph structures and 
 the relations play little role in.
Thus,
we claim that GraIL fails to learn the relational dependency in the ill-defined enclosing subgraph.

We also change the aggregation function in RED-GNN from \eqref{eq:aggregator}
to 
\[\bm h_{e_o}^\ell(e_q, r_q) = \delta\Big(\bm W^\ell 
\cdot \sum\nolimits_{(e_s, r, e_o)\in{\hat{\mathcal E}}_{e_q}^{\ell}} \alpha_{e_s,r_q}^\ell (\bm h_{e_s}^{\ell-1}(e_q, r_q)+\bm v^\ell)
\Big), \]
where $\bm v^\ell\in\mathbb R^d$ is shared in the same layer,
to remove the relation on edges.
We denote this variant as RED-no rel.
Note that RED-no rel is different from the variant 
Attn-w.o.-$r_q$ in Table~\ref{tab:abla}.
As in Table~\ref{tab:homo},
the performance drops dramatically.
RED-GNN mainly relies on the relational dependencies in edges
as evidence for reasoning.

\begin{table}[ht]
	\centering
	\caption{Comparison of GraIL with and without relation information. Evaluated by MRR.}
	\label{tab:homo}
	\vspace{-8px}
	\setlength\tabcolsep{3.5pt}
	\begin{tabular}{c|c|c|c}
		\toprule
		methods      & WN18RR (V1) & FB15k-237 (V1) & NELL-995 (V1) \\ \midrule
		GraIL no rel  &    0.620    &     0.255      &     0.475     \\
		GraIL       &    0.627    &     0.279      &     0.481     \\ \midrule
		RED-no rel &    0.557    &     0.198      &     0.384     \\
		RED-GNN      &    0.701    &     0.369      &     0.637     \\ \bottomrule
	\end{tabular}
\end{table}

\section{Proofs}
\label{app:proofs}

\subsection{Proposition~\ref{pr:eqneighbor}}

We prove this proposition from the bi-directions
such that 
$\cup_{e_a\in\mathcal V}\mathcal E_{e_q, e_a|L}^\ell$ 
$\subseteq\hat{\mathcal E}_{e_q}^\ell$
and
$\hat{\mathcal E}_{e_q}^\ell \subseteq \cup_{e_a\in\mathcal V}\mathcal E_{e_q, e_a|L}^\ell$.

\begin{proof}
	\vspace{-5px}
	First, we prove that $\cup_{e_a\in\mathcal V}\mathcal E_{e_q, e_a|L}^\ell\subseteq\hat{\mathcal E}_{e_q}^\ell$.
	
	As in step-7 of Algorithm~\ref{alg:simple},
	$\mathcal E_{e_q, e_a|L}^\ell$ is defined as 
	$\mathcal E_{e_q, e_a|L}^\ell\!=\!\{(e_s, r, e_o)\in\hat{\mathcal E}_{e_q}^{\ell}| e_o\in\mathcal V_{e_q, e_a|L}^{\ell}\}$.
	Thus we have $\mathcal E_{e_q, e_a|L}^\ell\subseteq \hat{\mathcal E}_{e_q}^\ell$ and also
	$\cup_{e_a\in\mathcal V}\mathcal E_{e_q, e_a|L}^\ell\subseteq\hat{\mathcal E}_{e_q}^\ell$.
	
	Second, we prove that $\hat{\mathcal E}_{e_q}^\ell \subseteq \cup_{e_a\in\mathcal V}\mathcal E_{e_q, e_a|L}^\ell$.
	
	We use proof by contradiction here.
	Assume that $\hat{\mathcal E}_{e_q}^\ell \subseteq \cup_{e_a\in\mathcal V}\mathcal E_{e_q, e_a|L}^\ell$
	is wrong, then there must exist $(e_s, r, e_o)\in \hat{\mathcal E}_{e_q}^\ell$ and $(e_s, r, e_o)\notin\mathcal E_{e_q, e_a|L}^\ell:\;\forall e_a\in\mathcal V$.
	In other words,
	there exists an edge $(e_s, r, e_o)$ that can be visited in $\ell$ steps walking from $e_q$,
	but not in any r-digraph $(e_s, r, e_o)\notin\mathcal E_{e_q, e_a|L}^\ell$.
	Based on Def.~\ref{def:relgraph} of r-digraph, then,
	$(e_s, r, e_o)$ will be an edge that is $\ell$ steps away from $e_q$
	but does not belong to the $\ell$-th triple of any relational paths $e_q \!\!\rightarrow_{r^1} \!\!\cdot\!\!\rightarrow_{r^2}\!\! \cdots \!\!\rightarrow_{r^L} \!\!e_a$,
	this is impossible.
	Therefore,
	we have $\hat{\mathcal E}_{e_q}^\ell \subseteq \cup_{e_a\in\mathcal V}\mathcal E_{e_q, e_a|L}^\ell$.
	
	Based on above two directions,
	we prove that $\hat{\mathcal E}_{e_q}^\ell = \cup_{e_a\in\mathcal V}\mathcal E_{e_q, e_a|L}^\ell$.
	\vspace{-5px}
\end{proof}

\subsection{Proposition~\ref{pr:identical}}

Given a query triple $(e_q, r_q, e_a)$ and $L$, 
when $\mathcal G_{e_q, r_q|L}=\emptyset$,
it is obvious that $\bm h_{e_a}^L(e_q, r_q)=\bm 0$ in both Algorithm~\ref{alg:simple}
and Algorithm~\ref{alg:gnn}.
For $\mathcal G_{e_q, r_q|L}\neq\emptyset$,
we prove by induction.
We denote 
$\hat{\bm{h}}_{e}^{\ell}(e_q,r_q)$ as the representations learned by Algorithm~\ref{alg:simple}
and 
$\breve{\bm{h}}_{e}^{\ell}(e_q,r_q)$ as the representations learned by Algorithm~\ref{alg:gnn}.
Then,
we show that $\hat{\bm{h}}_{e}^{\ell}(e_q,r_q)=\breve{\bm{h}}_{e}^{\ell}(e_q,r_q)$ 
for all $\ell=1\dots L$ and $e\in V_{e_q, e_a|L}^\ell$.

\begin{proof}
	~~
	\begin{itemize}[leftmargin=*]
		\item 
		When $\ell=1$,
		$\hat{\bm{h}}_{e}^{1}(e_q,r_q) = \delta
		(\bm{W}^{1} \!\cdot\! \sum\nolimits_{(e_q, r, e)\in\mathcal E_{e_q, e_a|L}^1} \phi\big(\bm 0, \bm h_r^1)\big),$
		and 
		$\breve{\bm{h}}_{e}^{1}(e_q,r_q) = \delta
		(\bm{W}^{1} \!\cdot\! \sum\nolimits_{(e_q, r, e)\in\mathcal E_{e_q}^1} \phi\big(\bm 0, \bm h_r^1)\big)$,
		for $e\in V_{e_q, e_a|L}^1$.
		It is obvious that
		$\{(e_q, r, e)\in\mathcal E_{e_q, e_a|L}^1\}\!\equiv\! \{ (e_q, r, e)\!\in\!\mathcal E_{e_q}^1|e\!\in\! V_{e_q, e_a|L}^1\}$.
		Thus,
		we have 
		$\hat{\bm{h}}_{e}^{1}(e_q,r_q)\!=\!\breve{\bm{h}}_{e}^{1}(e_q,r_q)$ 
		for all $e\!\in\! V_{e_q, e_a|L}^1$.

		\item Assume that $\hat{\bm{h}}_{e}^{\ell-1}(e_q,r_q) = \breve{\bm{h}}_{e}^{\ell-1}(e_q,r_q)$ 
		for all $e\in V_{e_q, e_a|L}^{\ell-1}$,
		we prove that $\hat{\bm{h}}_{e}^{\ell}(e_q,r_q) = \breve{\bm{h}}_{e}^{\ell}(e_q,r_q)$ 
		for all $e\in V_{e_q, e_a|L}^{\ell}$.
		
		For Algorithm~\ref{alg:simple}, we have 
		\begin{equation}
		\vspace{-2px}
		\hat{\bm{h}}_{e}^{\ell}(e_q,r_q) = \delta
		(\bm{W}^{\ell} \!\cdot\! \sum\nolimits_{(e_s, r, e)\in\mathcal E_{e_q, e_a|L}^\ell} \phi\big(\hat{\bm h}^{\ell-1}_{e_s}\!(e_q,r_q), \bm h_r^\ell)\big). \label{eq:app:a1}
		\vspace{-2px}
		\end{equation}
		For Algorithm~\ref{alg:gnn}, we have 
		\begin{equation}
		\vspace{-2px}
		\breve{\bm{h}}_{e}^{\ell}\!(e_q, r_q) = \delta
		(\bm{W}^{\ell} \cdot \sum\nolimits_{(e_s, r, e)\in\hat{\mathcal E}_{e_q}^{\ell}} \phi\big(\breve{\bm h}^{\ell-1}_{e_s}\!(e_q,r_q), \bm h_r^\ell\big)).  \label{eq:app:a2}
		\vspace{-2px}
		\end{equation}
		As in step~7 of Algorithm~\ref{alg:simple},
		$\mathcal E_{e_q, e_a|L}^\ell\!=\!\{(e_s, r, e_o)\in\hat{\mathcal E}_{e_q}^{\ell}| e_o\in\mathcal V_{e_q, e_a|L}^{\ell}\}$.
		Then,
		for $e\in V_{e_q, e_a|L}^{\ell}$,
		the ranges of summation in \eqref{eq:app:a1} and \eqref{eq:app:a2} are the same,
		i.e.,
		$\{(e_s, r, e)\in\mathcal E_{e_q, e_a|L}^\ell\} \equiv \{ (e_s, r, e)\in\hat{\mathcal E}_{e_q}^{\ell}|e\in V_{e_q, e_a|L}^{\ell}\}$
		and $e_s$ here are all belonging to $V_{e_q, e_a|L}^{\ell-1}$.
		Hence,
		based on the assumption,
		we have $\hat{\bm{h}}_{e}^{\ell}(e_q,r_q) = \breve{\bm{h}}_{e}^{\ell}\!(e_q, r_q)$ or all
		$e\in V_{e_q, e_a|L}^{\ell}$.
	\end{itemize}
	By induction,
	we can have $\hat{\bm{h}}_{e}^{\ell}(e_q,r_q)=\breve{\bm{h}}_{e}^{\ell}(e_q,r_q)$ 
	for all $\ell=1\dots L$ and $e\in V_{e_q, e_a|L}^\ell$.
	Therefore,
	the representation  $\hat{\bm{h}}_{e_a}^{\ell}(e_q,r_q)$ and  $\breve{\bm{h}}_{e_a}^{\ell}(e_q,r_q)$
	learned by Algorithm~\ref{alg:simple} and Algorithm~\ref{alg:gnn}, respectively,
	are identical.
\end{proof}

\subsection{Theorem~\ref{theo:interp}}

\begin{proof}
	Based on Definition~\ref{def:relgraph},
	any set $\mathcal P$ 
	of relational paths $e_q\rightarrow_{r_i^1} \rightarrow_{r_i^2}\cdots \rightarrow_{r_i^L}e_a$
	are contained in the r-digraph $\mathcal G_{e_q, e_a|L}$.
	
	Denote $G_{\mathcal P}$ as the r-digraph constructed by $\mathcal P$.
	In each layer,
	the attention weight is computed as 
	\begin{align*}
	\alpha_{e_s,r,e_o|r_q}^{\ell} &= 
	\sigma\Big(
	(\bm w_{\alpha}^{\ell})^\top
	\text{ReLU}\big(\bm W_{\alpha}^{\ell} \cdot
	(\bm{h}_{e_s}^{\ell-1}(e_q, r_q)\oplus
	\bm{h}_r^{\ell}\oplus 
	\bm{h}_{r_q}^{\ell})\big)\Big) \\
	&= \text{MLP}^\ell(e_s, r, e_q, r_q).
	\end{align*}
	
	Then, 
	we prove that $\mathcal G_{\mathcal P}$ can be extracted
	from the $L$-th layer and recursively to the first layer.
	
	\begin{itemize}[leftmargin=*]
		\item  In the $L$-th layer,
		denote $\mathcal T_+^L$  as the set of the triples $(e_s, r^L_i, e_a)$,
		whose attention weights are $\alpha_{e_s,r^L_i,e_a|r_q}^{L}=MLP^L(e_s, r^L_i, r_q, e_q)$,
		in the $L$-th layer of $\mathcal G_{\mathcal P}$.
		Based on the universal approximation theorem~\cite{hornik1991approximation},
		there exists a set of parameters $\bm w_{\alpha}^{L}, \bm W_{\alpha}^{L}, \bm{h}_r^{L}$
		that can learn a decision boundary $\theta$ that
		$(e_s, r^L_i, e_a) \in \mathcal T_+^L$ if $\alpha_{e_s,r^L_i, r_q}^L >\theta$ 
		and otherwise $(e_s, r^L_i, e_a) \notin \mathcal T_+^L$.
		Then the $L$-th layer of $\mathcal G_{\mathcal P}$ can be extracted.
		
		\item 
		Similarly,
		denote $\mathcal T_+^{L-1}$  as the set of the triples $(e_s, r^{L-1}_i, e_o)$
		that connects with the remaining entities in the $L-1$-th layer.
		Then,
		there also exists a set of parameters $\bm w_{\alpha}^{L-1}, \bm W_{\alpha}^{L-1}, \bm{h}_r^{L-1}$
		that can learn a decision boundary $\theta$ that
		$(e_s, r^{L-1}_i, e_o) \in \mathcal T_+^{L-1}$ if $\alpha_{e_s,r^{L-1}_i, e_o|r_q}^{L-1} >\theta$ and otherwise not in.
		Besides,
		$\bm w_{\alpha}^{L-1}, \bm W_{\alpha}^{L-1}, \bm{h}_r^{L-1}$ and 
		$\bm w_{\alpha}^{L}, \bm W_{\alpha}^{L}, \bm{h}_r^{L}$ are independent with each other.
		Thus, we can extract the $L-1$-tfh layer of $\mathcal G_{\mathcal P}$.
		
		\item
		Finally,
		with recursive execution,
		$\mathcal G_{\mathcal P}$ can be extracted as the subgraph in $\mathcal G_{e_q,e_a|L}$
		with attention weights $\alpha_{e_s,r, e_o|r_q}^\ell >\theta$.
	\end{itemize}
\end{proof}

\end{document}